\pdfoutput=1

\documentclass[10pt,twocolumn,letterpaper]{article}

\usepackage{cvpr}              % To produce the CAMERA-READY version
\usepackage{amssymb}
\usepackage{booktabs}
\usepackage{makecell}
\usepackage{multirow}
\usepackage{colortbl}
\usepackage{amsfonts}
\usepackage{tabularx}
\usepackage[linesnumbered,ruled,vlined]{algorithm2e}
\usepackage[dvipsnames]{xcolor}
\definecolor{cvprblue}{rgb}{0.21,0.49,0.74}
\usepackage[pagebackref,breaklinks,colorlinks,citecolor=cvprblue]{hyperref}

\def\paperID{15667} % *** Enter the Paper ID here
\def\confName{CVPR}
\def\confYear{2024}

\title{Improving Out-of-Distribution Generalization in Graphs \\via Hierarchical Semantic Environments}
\author{Yinhua Piao\textsuperscript{\rm 1}, Sangseon Lee\textsuperscript{\rm 2}, Yijingxiu Lu\textsuperscript{\rm 1}, Sun Kim\textsuperscript{\rm 1,3,4}\\
\\
Department of Computer Science and Engineering, Seoul National University\textsuperscript{\rm 1}\\
Institute of Computer Technology, Seoul National University\textsuperscript{\rm 2}, 
AIGENDRUG Co., Ltd.\textsuperscript{\rm 3}\\ 
Interdisciplinary Program in Artificial Intelligence, Seoul National University\textsuperscript{\rm 4}\\ 
{\tt\small \{2018-27910, sangseon486, solanoon0113, sunkim.bioinfo\}@snu.ac.kr}
}

\begin{document}
\maketitle
\begin{abstract}
Out-of-distribution (OOD) generalization in the graph domain is challenging due to complex distribution shifts and a lack of environmental contexts. 
Recent methods attempt to enhance graph OOD generalization by generating \textbf{flat} environments.
However, such flat environments come with inherent limitations to capture more complex data distributions. 
Considering the DrugOOD dataset, which contains diverse training environments (e.g., scaffold, size, etc.), flat contexts cannot sufficiently address its high heterogeneity.
Thus, a new challenge is posed to generate more semantically enriched environments to enhance graph invariant learning for handling distribution shifts.
In this paper, we propose a novel approach to generate \textbf{hierarchical semantic environments} for each graph. 
Firstly, given an input graph, we explicitly extract variant subgraphs from the input graph to generate proxy predictions on local environments. Then, stochastic attention mechanisms are employed to re-extract the subgraphs for regenerating global environments in a hierarchical manner. 
In addition, we introduce a new learning objective that guides our model to learn the diversity of environments within the same hierarchy while maintaining consistency across different hierarchies. 
This approach enables our model to consider the relationships between environments and facilitates robust graph invariant learning. 
Extensive experiments on real-world graph data have demonstrated the effectiveness of our framework. Particularly, in the challenging dataset DrugOOD, our method achieves up to 1.29\% and 2.83\% improvement over the best baselines on IC50 and EC50 prediction tasks, respectively.
\end{abstract}

\section{Introduction}
\label{sec:intro}
\begin{figure}
    \centering
    \includegraphics[width=1.0\linewidth]{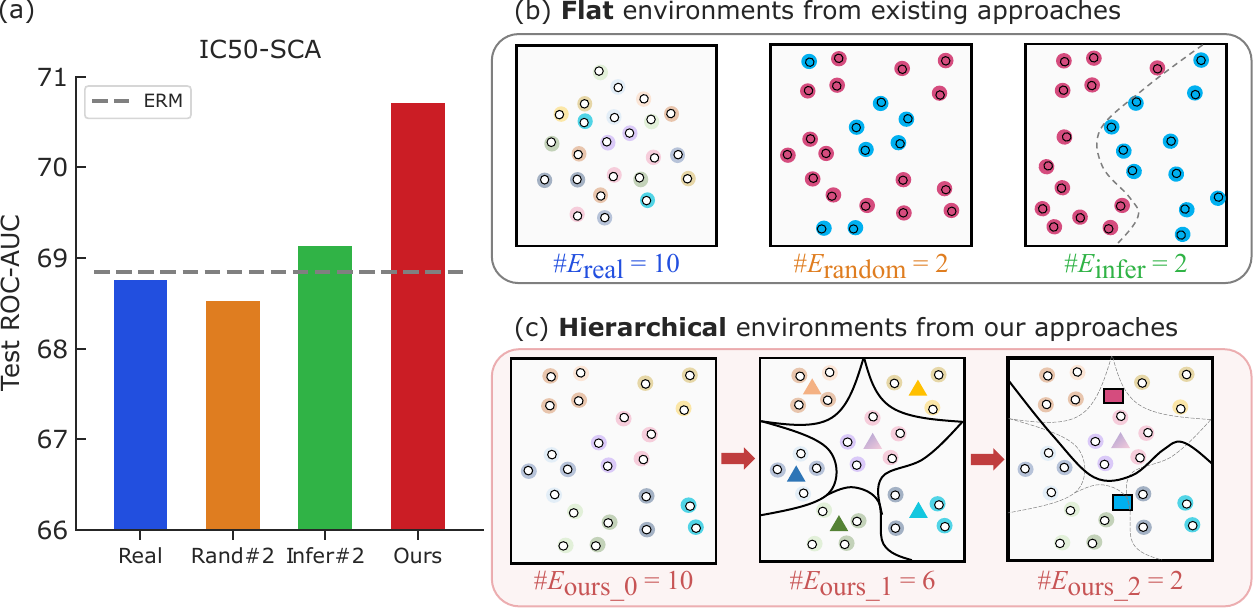}
    \caption{ (a) Results on $\textsc{IC50-sca}$ dataset from DrugOOD \cite{ji2023drugood}. 
    (b) Flat environments from existing approaches. (c) Hierarchical environments from our methods. For visualization, we set \#real environments as $10$.}
    \label{fig:heading}
\end{figure}
Graph-structured data is ubiquitous in real-world applications, from social networks to biological networks and chemical molecules \cite{jin2018junction, han2019gcn, hu2020open, zhang2022robust}. 
One notable advancement in this area is the emergence of Graph Neural Networks (GNNs). GNN-based models have pioneered end-to-end learning strategies to extract valuable information from graphs and have demonstrated remarkable success across various applications \cite{kipf2016semi, xu2018powerful, hamilton2017inductive}. 
However, the success of GNNs encounters challenges in out-of-distribution (OOD) scenarios primarily due to the intricate nature of graph distribution shifts \cite{david2020molecular, bevilacqua2021size, lim2022modeling}. Graph data, characterized by nodes, edges, and potential attributes, poses additional challenges compared to other domains such as natural language processing (NLP) or computer vision (CV).
Unlike those domains where context is often provided by sentences, paragraphs, or spatial information in images \cite{hsu2020generalized, chen2020graph, ming2022delving}, graph lacks a built-in contextual framework, making it inherently challenging to discern the relevance and context of individual graph elements in OOD scenarios \cite{zhou2020graph, gui2022good}.

Invariant Risk Minimization (IRM) \cite{arjovsky2019invariant} is a widely used strategy in the Euclidean domain, relying on the assumption that training data is sourced from distinct environments with varied data distributions. Motivated by the success of invariant learning in the Euclidean domain \cite{arjovsky2019invariant, ahuja2021invariance, krueger2021out, creager2021environment, huang2022environment, lin2022zin, liu2021heterogeneous, liu2021kernelized}, invariant learning methods for non-Euclidean graphs \cite{yang2022learning, li2022learning, chen2022learning, chen2024does, fan2022debiasing, li2022learning, liu2022graph} extend the concept to address the unique challenges posed by graph-structured data. 
Despite commendable progress in addressing OOD scenarios within the graph domain, existing studies have predominantly adhered to the assumption of a flat environmental situation \cite{chen2022invariance, liu2023flood}. 
For example, in the DrugOOD dataset, there are nearly 7000 diverse training environments in \textsc{ic50-sca} subset, including a wide variety of distinct substructures.
\cref{fig:heading} illustrates three existing approaches to acquire environments when performing invariant learning on this data: \textit{``Real''} means directly utilizing provided environments \cite{li2022learning}, \textit{``Rand\#2''} means randomly splitting samples into two environments \cite{chen2022learning, arjovsky2019invariant}, and\textit{ ``Infer\#2''} means inferring samples into two environments \cite{creager2021environment, huang2022environment}.
As shown in \cref{fig:heading}, all existing approaches exhibit poor or comparable performance to empirical risk minimization (ERM), which is consistent with the findings of Yang~\cite{yang2022learning}.
These phenomena indicate the shortcomings of flat environments: 
(1) Limited consideration of local environment similarity degrades performance in numerous environments (e.g., `\textit{Real}' case in \cref{fig:heading})
(2) Inferring from a small number of environments may fail to capture global environment similarities and interrelationships. (e.g., `\textit{Infer\#2}' case in \cref{fig:heading} )

Considering that graphs exhibit a hierarchical structure, with semantic information organized across various levels. This inherent hierarchy is fundamental to understanding structures and properties within graphs \cite{ying2018hierarchical}. In addition, inspired by the recent works \cite{lin2022zin, huang2022environment, ahuja2021invariance}, which show that the diversity of environments is recognized as the key to effectively 
handling graph OOD scenarios
, we aim to bridge the gap of flat environments while leveraging the inherent hierarchy of graphs. In this paper, we propose a hierarchical approach to generate semantic environments of each graph for effective graph invariant learning. 
Initially, we extract variant subgraphs from the input graph, enabling the generation of proxy predictions on local environments. Employing stochastic attention mechanisms, we iteratively re-extract subgraphs, building global environments hierarchically. 
To guide the robustness of hierarchical environment inference,
we introduce a hierarchical environment diversification loss which encourages our model to diversify environments within the same hierarchy while maintaining consistency across different hierarchical levels. This approach not only aids our model in considering relationships between environments but also strengthens graph invariant learning robustness.
By modeling relationships between different environments within a hierarchical framework, our approach acquires the rich source of environmental information embedded in the hierarchical structure of graphs. Extensive experiments demonstrate our approach to graph OOD classification datasets. Our contributions can be summarized as follows:
\begin{itemize}
    \item We propose a hierarchical approach to generate semantic environments for effective graph invariant learning. To the best of our knowledge, our proposed method is the first attempt to generate the environments in a hierarchical way in graph OOD generalization.
    \item We introduce a new learning objective that guides our model to learn the diversity of environments within the same hierarchy while maintaining consistency across different hierarchies.
    \item Extensive experiments have demonstrated our model yields significant improvements over various domains. In molecule graph benchmarks DrugOOD, our method achieves up to 1.29\% and 2.83\% higher ROC-AUC compared to SOTA graph invariant learning approaches. 

\end{itemize}

\section{Related Works}
\label{sec:relatedworks}
\paragraph{Out-of-Distribution Generalization}
In the Euclidean domain, such as computer vision, OOD generalization has been studied extensively using a variety of strategies, such as Invariant Risk Minimization (IRM) \cite{arjovsky2019invariant}, data augmentation \cite{zhang2017mixup}, and domain adaptation methods \cite{sun2016deep}.
Unlike deep neural networks with ERM, which suffer performance degradation under data distributional shifts, IRM-based studies \cite{krueger2021out, ahuja2021invariance} have shown powerful and robust performance when providing environmental information. 
Recently, a two-step optimization framework EIIL \cite{creager2021environment} has been proposed to train the invariant learning model without explicit environment labels. EIIL trains the environment inference model to learn and infer the environment labels in the first step, and in the second step, EIIL uses inferred environments to conduct invariant learning. Other recent approaches, such as ZIN \cite{lin2022zin}, HRM \cite{liu2021heterogeneous}, KerHRM\cite{liu2021kernelized}, and EDNIL \cite{huang2022environment} improve the OOD generalization by augmenting the environment to address its high heterogeneity. These augmentation strategies, such as the assistant model, clustering-based model, or effective diversification objective function all enhance the learning stage to infer diverse environments for effective invariant learning.
\paragraph{Graph Invariant Learning}
While general invariant learning methods are effective for the Euclidean domain, challenges arise when applying them to non-Euclidean domains like graph structures. In the realm of graph invariant learning, since graph data have an inherent nature of complex distribution shifts, often lacks explicit environment labels. To address this issue, methodologies have been widely developed and implemented. Approaches such as EERM \cite{wu2022handling}, LiSA \cite{yu2023mind}, and GREA \cite{liu2022graph} proposed to generate new environments, offering effective strategies to discover invariant patterns without environment label knowledge. While methods like MoleOOD \cite{yang2022learning} and GIL\cite{li2022learning} focus on the inference of environments, providing solutions to deal with challenging or unexplored environments. Another dimension of exploration involves learning with auxiliary assistance to ensure sufficient variation, with typical techniques found in CIGA \cite{chen2022learning}, GALA\cite{chen2024does}. Collectively, these innovative methods offer detailed strategies to navigate the challenges inherent in graph invariant learning. 
\section{Problem Definition and Preliminaries}
In \cref{subsec: problem definition}, we introduce notations about out-of-distribution generalization and the background knowledge of graph invariant learning.
In \cref{subsec: preliminaries}, we introduce environment inference to explain how recent research works on the graph dataset without environmental information. 

\subsection{Problem Definition}
\label{subsec: problem definition}
Following general settings of OOD generalization, we assume that the training datasets are collected from multiple training environments. Let $\mathcal{G}_{tr} = \{G^e\}_{e \in \texttt{supp}(\mathcal{E}_{tr})}$ with different graphs $G^e = \{ (G_i^e, y_i^e) \, | \, 1 \leq i \leq N^e \}$ be the training datasets, where $G^e$ represents the graphs drawn from environment $e$, and $\mathcal{E}_{tr}$ indicates the environment labels in the training datasets. 
We denote the environments $\mathcal{E}_{all}$ as all possible environment labels in the real world. Then, the test datasets contain graphs $\mathcal{G}_{test} = \{ G^{e'} \}$ with unknown environment labels $e'$, where $e' \in \texttt{supp}(\mathcal{E}_{all}) \setminus \texttt{supp}(\mathcal{E}_{tr})$. Our approach addresses out-of-distribution challenges in graph-level classification by incorporating invariant learning within environments derived from complex heterogeneous data hierarchically.

\subsection{Preliminaries}
\label{subsec: preliminaries}
\paragraph{Graph Invariant Learning.} Recently, several studies have identified the cause of performance deterioration in out-of-distribution (OOD) graphs, attributing it to the introduction of spurious features. 
It is widely acknowledged that spurious features exhibit sensitivity to environmental changes, often referred to as environmental factors. The incorporation of spurious features diminishes the significance of invariant features, resulting in a decline in OOD performance. As a result, recent research \cite{chen2022learning, chen2024does, li2022learning, yu2023mind} often involves graph decomposition to extract both variant subgraphs $G_v$ and invariant subgraphs $G_{inv}$ from the original graphs $G$, thereby facilitating the invariant GNN encoder $f$ \cite{kipf2016semi} to learn relationships between invariant features $h_{inv}$ and labels $y$.
Following the invariant risk minimization \cite{arjovsky2019invariant}, the encoder $f$ is trained with a regularization term enforcing jointly optimize the encoder $f$ in training environments $e$, which can generalize to unseen testing environments $e'$:
\begin{equation}
    \sum_{e\in{\texttt{supp}\{\mathcal{E}_{tr}\}}}{\mathcal{R}^e(f)+\lambda||\nabla{\mathcal{R}^e(f)}||^2},
\label{eq:1}
\end{equation}
where $\mathcal{R}^e(f)=\mathbb{E}^e_{G,y}[\mathcal{L}(f(G), y)]$
denotes the risk of $f$ in environment $e$, and $\mathcal{L}(\cdot,\cdot)$ denotes the loss function.

\paragraph{Environment Inference for GIL.}
In most cases, environmental information is difficult to obtain, especially in the context of graph generalization. While some datasets may provide metadata that could serve as environmental information, it is not reliable to fully use such information as the ground truth of the environment is not guaranteed.
Therefore, existing methods \cite{creager2021environment, huang2022environment, lin2022zin, liu2021heterogeneous, liu2021kernelized} often infer the environmental labels $\mathcal{E}_{infer}$ by using variant subgraphs $G_v$ that are related to the environment. A variant GNN encoder $f^e$ is used to learn the variant features $h_v$ from $G_v$ and infer the label $\hat{e}$ of the environment $e$. This approach allows joint optimization of environment inference and graph invariant learning through a learning process.

\section{Methods}
\begin{figure*}
  \centering
    \includegraphics[width=1\linewidth]{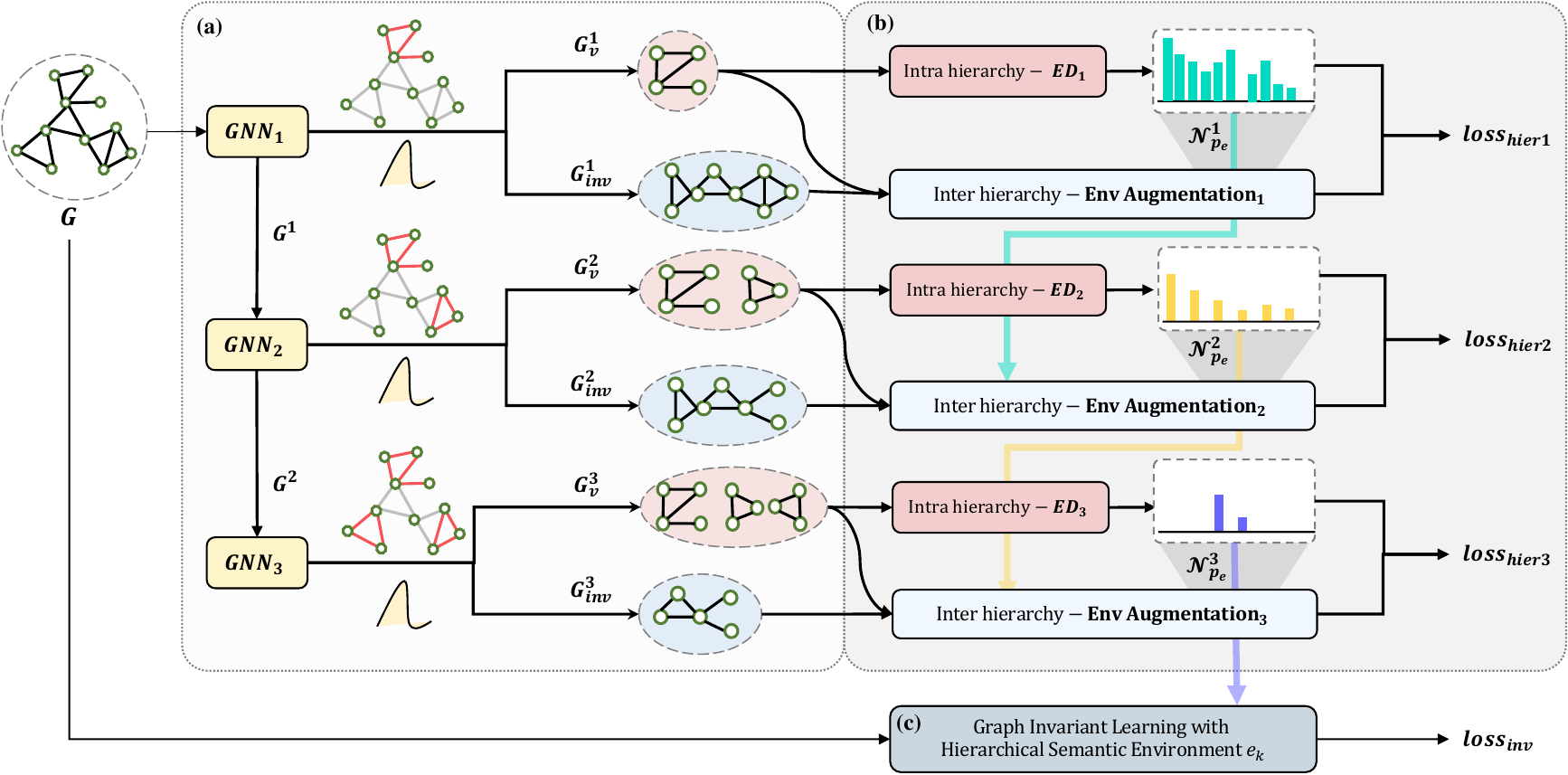}
    \label{fig:short-a}
  % \hfill
  \caption{Our Framework consists of (a) Hierarchical Stochastic Subgraph Generation in \cref{sec:HSSG}, (b) Hierarchical Semantic Environments in \cref{sec:HSE}, (c) Robust GIL with Hierarchical Semantic Environments in \cref{sec:GIL}. }
  \label{fig:mainfigure}
\end{figure*}
Our method consists of three components: hierarchical stochastic subgraph generation \cref{sec:HSSG}, hierarchical semantic environment inference \cref{sec:HSE}, and robust graph invariant learning (GIL) with inferred environments \cref{sec:GIL}. 
In \cref{sec:HSSG}, we generate invariant and variant subgraphs hierarchically for hierarchical semantic environment inference. In \cref{sec:HSE}, we learn hierarchical semantic environments using the proposed hierarchical loss. Finally, in \cref{sec:GIL}, we leverage the inferred hierarchical semantic environments to facilitate robust graph invariant learning, enabling us to uncover the invariant relationships between input graphs and their corresponding labels. These three steps together form a comprehensive approach for enhanced graph analysis and understanding.
\subsection{Hierarchical Stochastic Subgraph Generation}
\label{sec:HSSG}
Given a graph \(G = (V, E)\), where \(V\) represents the set of nodes and \(E\) represents the set of edges, the goal is to generate the variant subgraph \(G^{k}_v\) and invariant subgraph \(G^{k}_{inv}\) from the original graph \(G\) at each hierarchy \(k \in K\), where \(K\) denotes the number of environmental hierarchies. At each hierarchy \(k\), we employ a graph neural network denoted as \(\text{GNN}_{k}\) to update the hidden embedding \(h^{k}_i = \text{GNN}_{k}(h^{k-1}_i, A)\) for each node \(v_i \in V\). 
We use graph neural networks as \(\text{GNN}_{k}\) \cite{xu2018powerful} to aggregate information from neighboring nodes 
for node embedding updating.
In the process of learning variant and invariant subgraphs at each hierarchy, we define a probability distribution function \(S^{k}(\cdot)\) for edge selection as follows. We omit the hierarchy notations, while the computation of probability distribution is conducted within each hierarchy:
\begin{equation}
s_{ij} = S(h_{ij}) = \sigma(\text{MLP}([h_i, h_j]))    
\end{equation}
Here, \(s_{ij}\) is the probability of selecting edge \(e_{ij}\), \(\sigma\) is the sigmoid function, and \(\text{MLP}(h_i, h_j)\) is a multi-layer perceptron (MLP) that takes node embeddings \(v_i\) and \(v_j\) as inputs.

To introduce stochasticity in edge selection, we generate a sampler \(\mathbf{p}_{ij} \in \{0,1\}\) from the Bernoulli distribution \(\mathbf{p}_{ij} \sim \{\pi_1 := s_{ij}, \pi_0 := 1 - s_{ij}\}\). Gumbel-Softmax \cite{jang2016categorical} is then applied to obtain differentiable edge selection probabilities for each edge:

\begin{equation}
\hat{\mathbf{p}}_{ij} = \frac{\exp\left((\log{\pi_1} + \mathbf{g}_1)/\tau\right)}{\sum_{i\in\{0,1\}}{\exp\left((\log{\pi_i} + \mathbf{g}_i)/\tau\right)}}
\end{equation}
Here, \(\mathbf{g}_1\) and \(\mathbf{g}_0\) are i.i.d variables sampled from the Gumbel distribution, and \(\tau\) is a temperature parameter that controls the smoothness of the sampling process. A lower \(\tau\) leads to more categorical (hard) selections, while a higher \(\tau\) results in smoother (soft) probabilities. As \(\tau\) approaches 0, \(\hat{\mathbf{p}}_{ij}\) can be annealed to a categorical distribution.

To ensure hierarchical neighborhood consistency \cite{piao2022sparse} in the generation of stochastic subgraphs, we introduce an additional hierarchical neighbor masking function denoted as \(N^{k}\). This masking function compares the neighbor matrix from the previous hierarchy (\(k-1\)) with the currently selected neighbors, ensuring that nodes within the hierarchical neighborhood are consistently either included or excluded. The hierarchical neighbor mask \(N^{k}_{ij}\) can be defined as:
\begin{equation}
N^{k}_{ij} = N^{k-1}_{ij} + \boldsymbol{1}\{N^{k-1}_{ij} = 0 \text{ and } \mathbf{p}^{k}_{ij} > T\}
\end{equation}
where $T$ denotes the threshold for selecting the edges based on $\mathbf{p}^{(k)}_{ij}$, and $\boldsymbol{1}$ denotes indicator function.
Then we obtain the variant and invariant subgraph at each hierarchy $k$ as:
\begin{equation}
A^k_v \leftarrow A \odot N^k, A^k_{inv} \leftarrow A - A^k_{v}    
\end{equation}
where the adjacency matrix of the variant subgraph, denoted as \(A_v^{k}\), is derived by performing element-wise multiplication between the original edge matrix \(A\) and the hierarchical neighbor mask matrix \(N^{k}\). Conversely, the adjacency matrix of the invariant subgraph, denoted as \(A_{inv}^{k}\), is obtained by subtracting the edges included in the variant subgraph \(A_v^{k}\) from the original edge matrix \(A\) \cite{li2022learning,chen2022invariance, chen2024does}. 
By following this process iteratively across hierarchies, we construct a series of variant and invariant subgraphs that capture the hierarchical and structural nuances within the original graph, enabling a multi-level analysis of downstream tasks and analysis.

\subsection{Hierarchical Semantic Environments}
\label{sec:HSE}
Since the existing environmental information is not reliable or available, we design an environment inference model to assign the graphs to relatively reliable environments.
Given prior $p(e|G)$, we maximize the log-likelihood of $p(y|G)$ to obtain the posterior $p(e|G,y)$. As there is no solution to the true posterior, existing methods use clustering-based or variational distribution to approximate and infer the posterior. Inspired by a multi-class classification problem, we estimate the posterior using the softmax function as:
\[
P(\hat{e} \in \mathcal{E}_{tr} | X, Y) = \frac{\exp{(-l(f^e(\Phi(X),Y)))}}{\sum_{e'\in \texttt{supp}(\mathcal{E}_{tr})}{\exp(-l({f^{e'}(\Phi(X), Y)))}}}    
\]
where $f^e$ denotes neural network with $y^e$-class classifier and $\Phi$ indicates the variant subgraph generator.

\subsubsection{Intra-Hierarchy Environment Diversification}
In the previous section, we discussed the extraction of variant subgraphs $\{G^k_{v}| k \in K \}$ utilizing GNN encoders in conjunction with corresponding stochastic masking $\{p_{ij}^k | k \in K\}$ at each hierarchy $k$. 
To guide the learning process effectively, we incorporate an intra-hierarchy environment diversification loss with variant subgraph and estimated posteriors, denoted as $L_{ED}$:
\begin{equation}
L^k_{ED} = -\sum_i \max_{e^k}\log{P}(e^k|f^k({G^k_v}_{i}, y_i))
\end{equation}
Specifically, $L_{ED}$ is designed to guide the network $f^k$ to maximize the dependency between $e^k$ and $Y$ given the variant subgraph $G^k_v$ at each hierarchy $k$. 
By optimizing this loss, the network $f^k$ is ensured to distinguish relationships between the generated environments, fostering diversity within hierarchies.

\subsubsection{Inter-Hierarchy Environment Augmentation}
The goal of inter-hierarchy environment augmentation is to maximize the intra-class mutual information of the estimated invariant subgraphs while maximizing the intra-environment mutual information of the estimated variant subgraphs. Before introducing our objective function, a standard InfoNCE loss \cite{oord2018representation} can maximize the similarity between positive pairs and minimize the similarities between some randomly sampled negative pairs, which is defined as:
\[
\mathcal{L}_{\text{InfoNCE}}(z, \mathcal{N}_p, \tau) = 
-\log\left(\frac
{\exp(z \cdot \mathcal{N}_p(z) / \tau)}
{\sum_{n \in \mathcal{N}(z)}\exp(z\cdot n / \tau)}\right)
\]
where $\mathcal{N}_p(z)$ denotes positive samples for $z$, and $\mathcal{N}(z)$ denotes the batch samples for $z$, and $\tau$ denotes a temperature parameter.
To ensure the reliability and diversification of learned hierarchical environments, we introduce two inter-hierarchy semantic invariants: label-invariant and neighborhood-invariant. Inspired by \cite{chen2022invariance, chen2024does}, we introduce two contrastive learning losses: (1) Environment-based Contrastive learning loss $\mathcal{L}_{\text{EnvCon}}$ and (2) Label-based Contrastive learning loss $\mathcal{L}_{\text{LabelCon}}$. 
For environment-based contrastive learning, we define an environment-based neighborhood function $\mathcal{N}^k_{p_e}$ that constructs the positive set for the anchor sample, where samples have the same inferred environment. 
To preserve neighborhood consistency across hierarchies, the hierarchical neighborhood function $\mathcal{N}^{k}_{p_e}$ can be defined as :
\begin{equation}
\mathcal{N}^{k}_{p_e}(z) = \mathcal{N}^{k-1}_{p_e}(z) \cup \{z_i \mid e^k_i=e^k_z, z_i\in \mathcal{N}(z)\}
\end{equation}
Using $\mathcal{N}^{k}_{p_e}$, we formulate the environment consistency loss $\mathcal{L}_{\text{EnvCon}}$ as follows: 
\begin{equation}
\mathcal{L}^k_{\text{EnvCon}} = \mathbb{E}_{G^k_v\sim{p_d}}\left[\mathcal{L}_{\text{InfoNCE}}(z^k_v, \mathcal{N}^{k}_{p_e}(z^k_v), \tau)\right]
\end{equation}
where $p_d$ denotes the batch of data, and $z^k_v$ denotes the embedding of variant subgraph  $G^k_v$ from a multi-layer perception $\text{MLP}$ at $k$-th hierarchy. Environment consistency loss $\mathcal{L}^k_{\text{EnvCon}}$ attracts the variant subgraphs with the same environment labels and pushes the variant subgraphs with different environments, which can augment the power of learning the sufficiency of the environment.

For label-based contrastive loss, we define a label-based neighborhood function that represents the positive samples given ground truth label $y$, which can be defined as $\mathcal{N}_{p_y}(z) = \{z_i \mid y_i=y_z, z_i\in \mathcal{N}(z)\}$.
To measure the label consistency loss, we formulate $\mathcal{L}_{\text{LabelCon}}$ as follows:
\begin{equation}
\mathcal{L}^k_{\text{LabelCon}} = \mathbb{E}_{G^k_{\text{inv}}\sim{p_d}}\left[\mathcal{L}_{\text{InfoNCE}}(z^k_{\text{inv}}, \mathcal{N}_{p_y}(z^k_{\text{inv}}), \tau)\right]
\end{equation}
where $z_{\text{inv}}$ denotes the embedding of invariant subgraph $G^k_{\text{inv}}$ from MLP at $k$-th hierarchy. 
\subsubsection{Overall Objective Function}
Considering the diversity of intra-hierarchy and consistency of inter-hierarchy, we define a total loss of each hierarchy $k$ as follows: 
\begin{equation}
\mathcal{L}_{\text{hier}_k} =\mathcal{L}^k_{\text{ED}} + \alpha \cdot \mathcal{L}^k_{\text{EnvCon}} + \beta \cdot \mathcal{L}^k_{\text{LabelCon}}
\end{equation}
where $\alpha$ and $\beta$ denote coefficient parameters. And the overall objective loss can be defined as $\mathcal{L}_{\text{HEI}} = \sum_{k=1}^{K}\mathcal{L}_{\text{hier}_k}$
This formulation ensures a unified optimization objective, allowing the model to learn invariant subgraphs that capture both label and environment information effectively.

\subsection{Robust GIL with Hierarchical Semantic Environments}
\label{sec:GIL}
After the hierarchical learning stage of environment generation, we extract the generated environment $e^{K-1}$ from the last hierarchy $K-1$.  Our objective is to minimize the invariant risk given by \cref{eq:1}, employing the hierarchically generated semantic environments $e^{K-1}$. The objective function to calculate $\mathcal{L}_{inv}$ is as follows:
\begin{equation}
    % \begin{split}
        \min_{f} \mathcal{L}_{cls}(f) + \nabla_{\hat{e}}(\mathcal{L}_{cls}(f)) \text{ s.t. } \hat{e} = \arg \min_{e} \mathcal{L}_{\text{HEI}}
    % \end{split}
\end{equation}
This equation represents an optimization problem where the goal is to find a function \(f\) that minimizes the sum of two terms. The first term \(\mathcal{L}_{cls}(f)\) corresponds to the cross entropy loss of the classification task. The second term \(\nabla_{\hat{e}}(\mathcal{L}_{cls}(f))\) represents the gradient with respect to \(\hat{e}\), which is a specific environment.
In essence, the objective is to discover a function \(f\) that performs well on a classification task while maintaining stability in performance across different environments, adhering to the specified environmental minimization condition.

\begin{table*}[t]
\centering
\begin{tabular}{lllllllllll}
\toprule
\textbf{\textsc{methods}} & \textbf{\textsc{ic50-assay}} & \textbf{\textsc{ic50-sca}} & \textbf{\textsc{ic50-size}} & \textbf{\textsc{ec50-assay}} & \textbf{\textsc{ec50-sca}} & \textbf{\textsc{ec50-size}} \\ 
\textbf{\textsc{\#env}} & 311 & 6881 & 190 & 47 & 850 & 167 \\ 
\midrule
\textsc{erm}\cite{vapnik1991principles} & 71.79$\pm$0.27 & 68.85$\pm$0.62 & 66.70$\pm$1.08 & 76.42$\pm$1.59 & 64.56$\pm$1.25 & 62.79$\pm$1.15 & \\
% \midrule 
\textsc{irm}\cite{arjovsky2019invariant}     &  72.12$\pm$0.49 & 68.69$\pm$0.65 & 66.54$\pm$0.42 & 76.51$\pm$1.89 & 64.82$\pm$0.55 & 63.23$\pm$0.56  \\
\textsc{v-rex}\cite{krueger2021out}   &  72.05$\pm$1.25 & 68.92$\pm$0.98 & 66.33$\pm$0.74 & 76.73$\pm$2.26 & 62.83$\pm$1.20 & 59.27$\pm$1.65  \\
\textsc{eiil}\cite{creager2021environment}    &  72.60$\pm$0.47 & 68.45$\pm$0.53 & 66.38$\pm$0.66 & 76.96$\pm$0.25 & 64.95$\pm$1.12 & 62.65$\pm$1.88  \\
\textsc{ib-irm}\cite{ahuja2021invariance}  &  72.50$\pm$0.49 & 68.50$\pm$0.40 & 66.64$\pm$0.28 & 76.72$\pm$0.98 & 64.43$\pm$0.85 & 64.10$\pm$0.61  \\
\midrule 
\textsc{grea} \cite{liu2022graph}    &  72.77$\pm$1.25 & 68.33$\pm$0.32 & 66.16$\pm$0.46 & 72.44$\pm$2.55 & \underline{67.98$\pm$1.00} & 63.93$\pm$3.01  \\
% \textsc{lisa}     &  63.50$\pm$4.60 & 58.30$\pm$4.70 & 59.60$\pm$4.70 & 68.10$\pm$2.60 & 64.60$\pm$1.70 & 61.20$\pm$1.90  \\ 
% \midrule
% \textsc{disc}    &  -              & -              & -              & 65.40$\pm$5.34 & 54.97$\pm$3.86 & 56.97$\pm$2.56 \\
% \textsc{cigav1}   &  72.50$\pm$0.70 & 68.64$\pm$0.36 & 66.46$\pm$0.49 & 78.46$\pm$0.45 & 66.05$\pm$1.29 & 66.01$\pm$0.84 \\
\textsc{cigav1} \cite{chen2022learning}  &  72.71$\pm$0.52 & 69.04$\pm$0.86 & 67.24$\pm$0.88 & 78.46$\pm$0.45 & 66.05$\pm$1.29 & \underline{66.01$\pm$0.84} \\
%\textsc{cigav2}   & 71.66$\pm$1.13 & 68.08$\pm$1.30 & 67.16$\pm$0.42 & 76.96$\pm$3.06 & 65.96$\pm$3.06 & 63.65$\pm$0.48 \\
\textsc{cigav2} \cite{chen2022learning}& \underline{73.17$\pm$0.39} & \underline{69.70$\pm$0.27} & \underline{67.78$\pm$0.76} & - & - & - \\
% \midrule
% \textsc{gil}     &  -              & -              & -              & 72.13$\pm$4.70 & 63.05$\pm$1.04 & 62.08$\pm$1.06 \\
\textsc{moleood} \cite{yang2022learning} &  71.38$\pm$0.68 & 68.02$\pm$0.55 & 66.51$\pm$0.55 & 73.25$\pm$1.24 & 66.69$\pm$0.34 & 65.09$\pm$0.90  \\
\textsc{gala} \cite{chen2024does}    &  - & - & - & \underline{79.24$\pm$1.36} & 66.00$\pm$1.86 & \underline{66.01$\pm$0.84} \\
\midrule
\rowcolor{Cyan!10}\textsc{ours} &  \textbf{74.01$\pm$0.11} & \textbf{70.72$\pm$0.30} & \textbf{68.64$\pm$0.23} & \textbf{80.82$\pm$0.21} & \textbf{69.73$\pm$0.21} & \textbf{66.87$\pm$0.38} \\ 
\bottomrule
\end{tabular}
\caption{Test ROC-AUC of various models on DrugOOD benchmark datasets. The mean $\pm$ standard deviation of all models is reported as an average of 5 executions of each model. The best methods are highlighted in \textbf{bold} and the second best methods are \underline{underlined}.
}
\label{table_drugood}
\end{table*}

\section{Experiments}
\paragraph{Datasets.}
Extensive experiments were conducted on real-world graph data from diverse domains.
\begin{itemize}
    \item \textbf{CMNIST-75sp.} The task is to classify each graph that is converted from an image in the ColoredMNIST dataset \cite{arjovsky2019invariant} into the corresponding handwritten digit using the superpixels algorithm \cite{knyazev2019understanding}. Distribution shifts exist on node attributes by adding random noises in the testing set.
    \item \textbf{Graph-SST datasets.} We utilize sentiment graph data from SST5 and SST-Twitter datasets from \cite{yuan2022explainability}. For the Graph-SST5 dataset, graphs are split into different sets based on averaged node degrees to create distribution shifts. For the Graph-Twitter dataset, we invert the split order to assess the out-of-distribution generalization capability of GNNs trained on large-degree graphs to smaller ones.
    \item \textbf{DrugOOD.} We use six datasets in DrugOOD \cite{ji2023drugood} that are provided with manually specified environment labels. DrugOOD provides more diverse splitting indicators, including assay, scaffold, and size. To comprehensively evaluate the performance of our method under different environment definitions, we adopt these three different splitting schemes on categories IC50 and EC50 provided in DrugOOD. Then we obtain six datasets, \textbf{EC50-*} and \textbf{IC50-*}, where the suffix * specifies the splitting scheme i.e. \textsc{IC50/EC50-assay/scaffold/size}.
\end{itemize}

\paragraph{Baselines.} We comprehensively compare our methods with the following categories of baselines: (1) ERM denotes supervised learning with empirical risk minimization \cite{vapnik1991principles}. (2) Euclidean OOD methods: We compare with SOTA invariant learning methods from the Euclidean regime, including IRM \cite{arjovsky2019invariant}, V-REX \cite{krueger2021out}, EIIL \cite{creager2021environment}, IB-IRM  \cite{ahuja2021invariance}. (3) Graph OOD methods: We also compare with SOTA invariant learning methods from the graph regime. Graph OOD methods can be further split into three groups: environment generation-based baseline methods including GREA \cite{liu2022graph} and LiSA \cite{yu2023mind}, environment augmentation-based baseline methods including DisC  \cite{fan2022debiasing} and CIGA \cite{chen2022learning} and environment inference-based baseline methods including GIL \cite{li2022learning}, MoleOOD \cite{yang2022learning}, and GALA \cite{chen2024does}.

\paragraph{Environmental Setup.}
All methods use the same GIN backbone \cite{xu2018powerful} and the same optimization protocol for fair comparisons. Each of the methods is configured using the same parameters reported in the original paper or selected by grid search. For a fair comparison, we use the same embedding size for all methods. We tune the hyperparameters following the common practice. All details are given in Appendix. We report the ROC-AUC score in the DrugOOD dataset and the accuracy score for the rest of the datasets.

\subsection{Performance Comparison}
We first provide a detailed report on the DrugOOD benchmark dataset in \cref{table_drugood}. We conduct IC50 and EC50 predictions with different split settings. As shown in \cref{table_drugood} and mentioned in \cref{sec:intro}, the DrugOOD dataset includes molecular datasets with complex distributions, such as assay, scaffold, and size split with various numbers of environments.
Our approach consistently outperforms existing methods, demonstrating the importance of hierarchical environment learning in addressing complex drug OOD generalization applications. 
As shown in \cref{table_drugood}, compared to Empirical Risk Minimization (ERM), Euclidean-based invariant learning methods, such as IRM \cite{arjovsky2019invariant}, V-REX \cite{krueger2021out}, EIIL \cite{creager2021environment}, etc, show comparable or even degraded performance in most experimental settings. This suggests that directly applying general invariant learning struggles to handle shifts in graph distributions. In contrast, graph-based OOD methods exhibit better performance, as they learn invariant subgraph patterns for graph OOD generalization.

In addition, in the performance comparison of graph-based OOD methods,  the MoleOOD \cite{yang2022learning} pointed out that simple ERM sometimes outperforms several existing methods when faced with a large number of provided environments in graph OOD datasets, which is aligned with the results of \cref{table_drugood}. 
For example, \textsc{ic50-sca} dataset is provided with 6881 environments, where 6881 scaffolds or substructures construct a vast amount of environmental information for the dataset. In \textsc{ic50-sca} dataset, GREA and MoleOOD show poor performance compared to ERM, as shown in \cref{table_drugood}.
We analyze this phenomenon for two potential reasons: (1) The first reason is that the provided environments are overly fine-grained, thereby similar substructure environments in the training and testing sets. ERM can capture similar substructures to learn similar data distributions, showing better performance. 
(2) The second reason is that GREA and MoleOOD only generate a small number of environments, 
which overly simplifies the environment and neglects the relationships between environments. This limits their ability to learn heterogeneous and interrelated environments, resulting in noticeable performance degradation.

Our method performs a hierarchical approach to learn relationships between redundant environments and maximize the diversity of environments. Leveraging the high-level semantic environments from the last hierarchy, we minimize invariant prediction risk, thereby achieving better graph OOD generalization.
In Appendix, we report performance comparisons showing that our performance is better or comparable to existing methods in general domains. 

\subsection{Effect of Hierarchical Semantic Environments}
In this section, similar to existing environment inference-based works for OOD generalization \cite{chen2024does, creager2021environment}, we first analyze the role of environment inference, and then we discuss the effects of hierarchical environments learned by our model, showing the necessity of learning the hierarchical environments for graph invariant learning.
\begin{table}[t]
\centering
\begin{tabular}{l|lllllll}
\toprule
\textbf{\textsc{methods}} & \textbf{\textsc{ic50-sca}} & \textbf{\textsc{ic50-size}}\\ \midrule
w/ $\text{env}_\text{\#non-infer(rand)=2}$ & 68.54$\pm$0.64 & 67.63$\pm$0.33 \\
w/ $\text{env}_\text{\#non-infer=real}$ & 68.77$\pm$0.72 & 67.60$\pm$0.32 \\
\midrule
w/ $\text{env}_{\#\text{infer}=\text{2}}$ & 69.14$\pm$0.80 & 67.55$\pm$0.34 \\
w/ $\text{env}_{\#\text{infer}=\text{real}}$ & 69.08$\pm$0.64 & 67.74$\pm$0.13 \\
\midrule
\textbf{w/ $\text{env}_{\#\text{hier-infer}}$ (\textsc{ours})}  & \textbf{70.72$\pm$0.30} & \textbf{68.64$\pm$0.23}\\ 

\bottomrule
\end{tabular}
\caption{The ablation study on \textsc{ic50-sca} and \textsc{ic50-size} datasets.}
\label{table_ablation}
\end{table}

\begin{table}[t]
\centering
\begin{tabular}{l|ll}
\toprule
\textbf{\textsc{Configurations}} & \textbf{\textsc{ic50-sca}} & \textbf{\textsc{ic50-size}} \\ \midrule
\#real = [\#$e_{p}$]             & 69.08$\pm$0.64 & 67.60$\pm$0.32\\
\#env = [5]                 & 69.35$\pm$0.67 & 67.70$\pm$0.50\\
\#env = [2]                 & 69.14$\pm$0.80 & 67.73$\pm$0.64\\
% \midrule
% \#real - \#real/2       & -$\pm$- \\
% \#real - 5              & 69.25$\pm$0.12 \\
% \#real - 2              & 68.80$\pm$0.50 \\
\midrule
\#env=[\#$e_{p}$ $\to$ \#$e_{p}/2$ $\to$ 5]   & 70.12$\pm$0.14 & 68.62$\pm$0.34 \\
\textbf{\#env=[\#$e_{p}$ $\to$ \#$e_{p}/2$ $\to$ 2]}   & \textbf{70.72$\pm$0.30} & \textbf{68.64$\pm$0.23} \\
% \midrule
% \rowcolor{Cyan!10} \textsc{ours}           & \textbf{70.72$\pm$0.30}\\ 
\bottomrule
\end{tabular}
\caption{Sensitivity analysis on generated environments. \#$e_{p}$ denotes the number of provided environments in datasets. }
\label{table_sensitive}
\end{table}

\paragraph{Environment Inference.} 
As shown in \cref{table_ablation}, we analyze the role of environment inference by comparing the direct usage of \textit{non-inferred} and \textit{inferred} environments. 
Direct usage of the \textit{non-inferred} environments can be divided into two scenarios: (1) When environments are unavailable, existing methods randomly assign two environments for graph invariant learning, which is shown as `w/ $\text{env}_{\text{\#non-infer (rand)}}$' from \cref{table_ablation}. 
(2) DrugOOD dataset provides environmental information, e.g., scaffold information in \textsc{ic50-sca} dataset. As shown in \cref{table_drugood}, we can know the number of environments for both dataset \textsc{ic50-sca} and dataset \textsc{ic50-size}. In such cases, we use environments for invariant learning, as illustrated by the `w/ $\text{env}_{\text{\#non-infer (real)}}$' in \cref{table_ablation}.

Secondly, we consider the usage of \textit{inferred environments} for invariant learning, as indicated in \cref{table_ablation}. Through the model of environment inference, we directly predict the flat environment and use it in invariant learning, as illustrated by `w/ $\text{env}_{\#\text{infer=*}}$'. 
From \cref{table_ablation}, it can be observed that the model with inferred environments exhibits better performance, indicating that learned environments from data are more effective than direct usage of \textit{non-inferred} environments. Moreover, in dataset \textsc{ic50-sca}, the effects of inference are more pronounced, emphasizing the necessity of performing environment inference operations in situations with complex real-world environments.

\paragraph{Hierarchical Environment Inference.}
We further analyze the effects of \textit{hierarchical} environments by comparing them with \textit{flat} environments. Our model adopts a hierarchical approach to learning environmental information, regulating semantic environmental content between hierarchies through both intra-hierarchy and inter-hierarchy mechanisms. Upon comparing experimental results in \textsc{ic50-sca} and \textsc{ic50-size} datasets, as shown in \cref{table_ablation}, our hierarchical model consistently outperforms invariant learning models with flat environments. This indicates that in real-world datasets, the environmental factors influencing label predictions are complex and interdependent. Consequently, investigating graph invariant learning with hierarchical environment inference is essential for attaining more sophisticated and effective in graph OOD generalization.

\subsection{Sensitive analysis on hierarchy}
We also conduct sensitive analysis on the number of hierarchies and the number of environments at each hierarchy. 
Specifically, in hierarchical settings, we start from the number of environments as the number of environments provided in the first hierarchy for a fair comparison. 
By comparing the results shown in \cref{table_sensitive}, we find inferred environments with hierarchical settings help OOD generalization in the graph domain. 
Inferring flat environments is worse than environment inference with three hierarchies. In addition, inferring a large number of environments $\text{\#}e_p$ in a single hierarchy shows poor performance compared to inferring a small number of environments. This is because the model with a large number of environments learns too local and redundant to capture the relationships between environments. 
However, in a hierarchical setting, even if inference starts from a large number of environments to a small number of environments, different hierarchies can capture the local relationships among a huge number of environments and dynamically maximize the diversity of inferred high-level environments. Therefore, despite the final number of inferred environments is still small, these environments learn the semantic information from complex data distribution hierarchically and can improve the Graph OOD generalization using the high-level semantic environments.

\subsection{Discussions}
\begin{figure}
    \centering
    \includegraphics[width=0.7\linewidth]{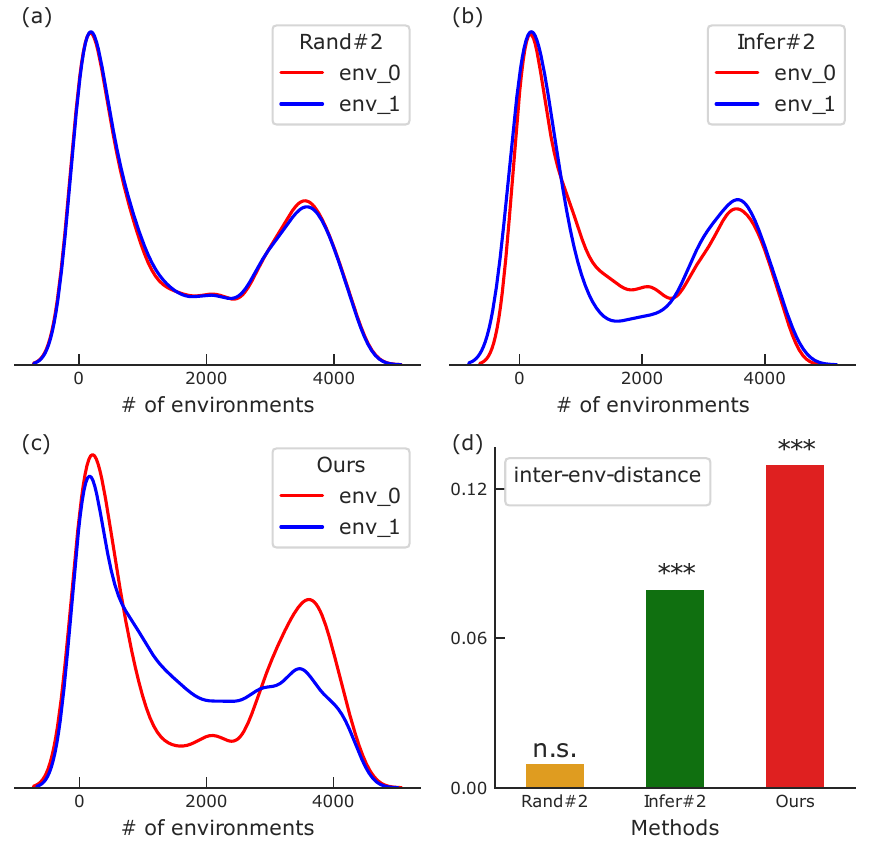}
    \caption{Discussions on the diversity of generated environments. We show distributions of two generated environments $env_0$ and $env_1$ for (a) random sampling methods, (b) flat environment inference methods, and (c) our hierarchical environment inference methods. (d) We employ the Kolmogorov-Smirnov test \cite{massey1951kolmogorov} to calculate the diversity of three methods.\vspace{-15pt}}
    \label{fig:discussion}
\end{figure}

We discuss the diversity of environments generated by different methods as shown in \cref{fig:discussion}. we compare two environment assignment methods with our method,  which are random sampling-based method \textit{rand\#2}, and flat environment inference-based method \textit{infer\#2}, respectively. For a fair comparison, we set two environments for both methods and set the hierarchical environment as $[6881 \to 3440 \to 2]$ for our method using the \textsc{ic50-sca} dataset from DrugOOD. 

As mentioned in \cite{creager2021environment}, the diversity of environments is important to obtain effective IRM models, and large discrepancy of spurious correlations between environments benefits the IRM model.
We measure the diversity of environments by calculating the distance of distributions among the acquired environments for each method. 
More specifically, we measure the Kolomogorov-Smirnov (K-S) statistic \cite{massey1951kolmogorov} as a distance between the two inferred scaffold distributions for each method, respectively. The K-S test can capture the dissimilarity between cumulative distribution functions, making it suitable for our analysis.
As shown in \cref{fig:discussion}, the two environments generated by \textit{rand\#2} methods are almost the same since they assign the sample to two environments using random sampling, showing the lowest diversity of environments. \textit{infer\#2} shows better diversity of generated environment distributions with significant p-values ($p=1.86e-26$).
Our method generates more significantly diverse environment distributions ($p=4.52e-73$) compared to other methods, surpassing them in \textit{inter-env-distance} by a large margin. Moreover, the results are aligned with \cref{fig:heading}, our methods outperform other methods in \textsc{ic50-sca} dataset, indicating that diverse environments can give a clear indication of distribution shifts, therefore IRM can easily identify and eliminate variant features.
\section{Conclusion}
In our research, we take on the formidable challenge of improving out-of-distribution (OOD) generalization in the graph domain. We observe that recent works often overlook the crucial aspect of studying hierarchical environments in graph invariant learning. To address this gap, we introduce a novel method that generates hierarchical dynamic environments for each graph.
Our approach involves hierarchical stochastic subgraph generation, hierarchical environment inference, and a carefully designed learning objective. By incorporating graph invariant learning with inferred high-level environments, our model achieves meaningful and diverse environments within the same hierarchy and ensures consistency across different hierarchies. The effectiveness of our method is particularly pronounced in the DrugOOD dataset, shedding light on the potential for further exploration in hierarchical graph learning within OOD scenarios.
\vspace{-5pt}
\section{Acknowledgement}
\vspace{-5pt}
This work was supported by Bio \& Medical Technology Development Program of the NRF \& ICT(NRF-2022M3E5F3085677, RS-2023-00257479), RS-2023-00246586 and Institute of Information \& communications Technology Planning \& Evaluation (IITP) grant funded by the Korea government(MSIT) [RS-2021-II211343]

{
    \small
    \bibliographystyle{ieeenat_fullname}
    \bibliography{main}
}
\newpage
\section{Implementation Details}
Our proposed method contains three components: hierarchical stochastic subgraph generation (\cref{Appendix_sec: hssc}), hierarchical semantic environments (\cref{Appendix_sec: hse}), and graph invariant learning. In this section, we extend the details of our architectures. We provide a detailed description of our model in \cref{algo:1} at the end of this section.

\subsection{Hierarchical Stochastic Subgraph Generation}
\label{Appendix_sec: hssc}

% We first use GINs \cite{xu2018powerful} to encode the input graph. Different GINs are used in different hierarchies. 
% We use two-layer MLPs with the ReLU activation function as a scoring function for edge selection at each hierarchy. Likewise, the scoring functions have different weight parameters at different hierarchies.
% We generate a sampler from the Bernoulli distribution drawn from the calculated score of edges. Then we select the edges using the Gumbel-Softmax function with threshold $T$ and temperature $\tau$. We set a threshold of 0.5 to select edges and a temperature of 0.05 during the training stage. 
% The variant subgraphs are then extracted by selected edges at each hierarchy, and the invariant subgraphs are obtained by directly subtracting the selected edges from the original input graph.
% In addition, we preserve the selected edges in previous hierarchies for stable training. 

We employ graph isomorphism networks (GINs) \cite{xu2018powerful} for graph encoding, utilizing different GINs for different hierarchies. A two-layer multilayer perceptron (MLP) with ReLU activation functions serves as the scoring function for edge selection in each hierarchy, with different weight parameters.
A sampler is generated from a Bernoulli distribution based on the computed edge scores. Edge selection uses the Gumbel-Softmax function with a threshold $T$ and temperature $\tau$. During training, we set the temperature at 0.05 for edge selection and we conduct a grid search for the threshold.
Variant subgraphs are extracted by the selected edges at each hierarchy, while invariant subgraphs are obtained by directly subtracting the selected edges from the original input graph. Notably, we retain the selected edges from previous hierarchies for stable training.
\subsection{Hierarchical Semantic Environments}
\label{Appendix_sec: hse}

\subsubsection{Intra-Hierarchy Environment Diversification}

To predict environment labels in each hierarchy, we employ a simple neural network, denoted as $f^e$, for multi-classification, drawing inspiration from the work \cite{huang2022environment}. 
Following Eq. 6, the input of the neural network $f^e$ is formed by concatenating the hidden embedding of the variant subgraph and the one-hot encoding of the label. 
The output dimension of the neural network $f^e$ corresponds to the number of environment labels at each hierarchy. 
During inference, predicted environment labels at each hierarchy are obtained by the neural network's output.
\begin{figure}
    \centering
    \includegraphics[width=1.0\linewidth]{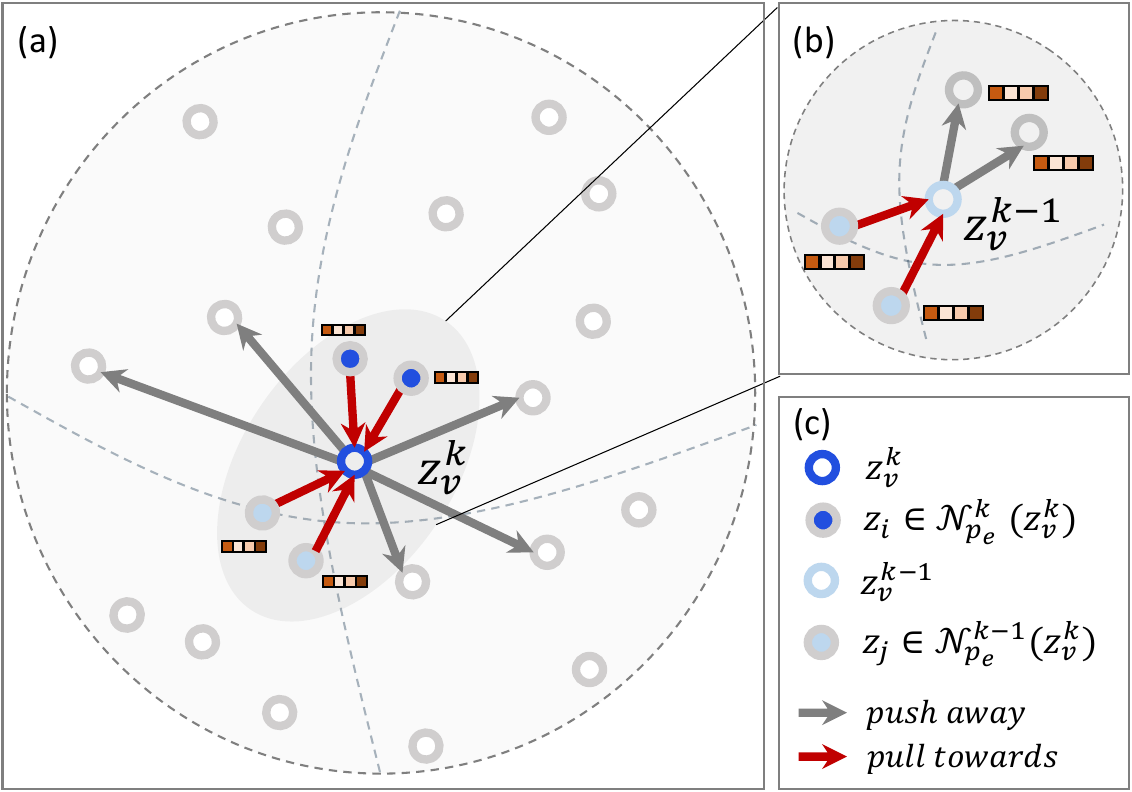}
    \caption{Illustration of objective $\mathcal{L_\text{EnvCon}}$ in Inter-Hierarchy Environment Augmentation. (a) We pull environment-based neighborhoods $\mathcal{N}^k_{p_e}(z^k_v)$ and $\mathcal{N}^{k-1}_{p_e}(z^k_v)$ toward anchor variant subgraph embedding $z^k_v$. (b) We show a simple illustration of anchor variant subgraph $z^{k-1}_v$ in the previous hierarchy $k-1$. (c) Notations of the illustration figure.}
    \label{fig:appendix-envcon}
\end{figure}

\subsubsection{Inter-Hierarchy Environment Augmentation}
% As highlighted by Huang \cite{huang2022environment}, multi-class classification often outperforms clustering-based methods. However, in hierarchical settings, we observe challenges in consistently assigning samples and their previous neighbors from previous hierarchies to the same class. To tackle this issue, we propose a contrastive objective that enhances environmental semantics in the latent embedding space by preserving neighborhood consistencies across hierarchies.

% Drawing inspiration from previous studies \cite{chen2024does, chen2022learning}, we employ contrastive learning on two dimensions: one based on the environment and the other on labels. In Figure \ref{fig:appendix-envcon}, we illustrate the $\mathcal{L}_{\text{EnvCon}}$ objective. For each anchor variant subgraph, positive neighborhoods encompass graphs with the same predicted environmental labels as the anchor graphs, including neighborhoods from the previous hierarchy. The $\mathcal{L}_{\text{LabelCon}}$ objective is computed similarly to $\mathcal{L}_{\text{EnvCon}}$, but the label-based neighborhoods of anchor subgraphs consistently remain the same. Notably, to distinguish embeddings between variant and invariant subgraphs, we employ two MLP layers with distinct parameters to embed each subgraph, respectively.

As emphasized by Huang \cite{huang2022environment}, multi-class classification often outperforms clustering-based methods. However, in hierarchical settings, we encounter challenges in consistently assigning samples and their neighbors from previous hierarchies to the same class. To address this issue, we propose a contrastive objective that enhances environmental semantics in the latent embedding space by preserving neighborhood consistencies across hierarchies.
Taking inspiration from prior studies \cite{chen2024does, chen2022learning}, we employ contrastive learning on two dimensions: one based on the environment and the other on labels. In \cref{fig:appendix-envcon}, we illustrate the objective $\mathcal{L}_{\text{EnvCon}}$. For each anchor variant subgraph, positive neighborhoods encompass graphs with the same predicted environmental labels as the anchor graphs, including neighborhoods from the previous hierarchy. 
The objective $\mathcal{L}_{\text{LabelCon}}$ is computed similarly to $\mathcal{L}_{\text{EnvCon}}$, but the label-based neighborhoods of anchor subgraphs consistently remain the same. Notably, to distinguish embeddings between variant and invariant subgraphs, we employ two MLP layers with distinct parameters to embed each subgraph.

% For example, $\{z_j \in \mathcal{N}^{k-1}(z^k_v)\}$ have the same predictive environment labels as $z^{k-1}_v$, thus they are the positive neighborhoods of anchor subgraph at hierarchy $k-1$. 

\begin{algorithm}[ht]
\caption{The training procedure.}
\KwData{Dataset $\mathcal{G}^{train}=\{(G_i, y_i)\}_{i=1}^{N_{train}}$, hierarchical subgraph generation module $s$, hierarchical environment inference module $f$, graph invariant learning module $F$. Number of hierarchies $K$, number of training epochs for hierarchical environment inference module $E_1$, number of training epochs for the graph invariant learning $E_2$, Batch size $B$.}
\KwResult{Final estimated model $F$}
\SetAlgoNlRelativeSize{0}
\BlankLine
\textbf{Initialize} the models $s, f, F$\;
\For{${p_1} \gets 1$ to $E_1$}{
    Sample a batch of data $\mathcal{B}$ from $\mathcal{G}_{\text{train}}$ with batch size $B$\;
    \For {$G_j$, $y_j \in \mathcal{B}$}{
        \For{$k \gets 1$ to $K$}{ 
            Obtain environment predictions $e^k_j$ from $f^k$;
            Find environment-based neighbors $\mathcal{N}^k_{pe}(G_j)$ with the same $e^k_j$ in $k$ and $k-1$-th layer\;
            Find label-based neighbors $\mathcal{N}_{py}(G_j)$ with the same $y_j$\;
            Compute loss $\mathcal{L}^k_{\text{hier}}(s^k, f^k;G_j)$ via Eq. 10\;
        }
        Compute loss $\mathcal{L}_{\text{HEI}}$ across all $K$ hierarchies\;
        Back propagate and optimize $s, f$\;
    }
}
Freeze the parameters of module $s, f$\;
\For{${p_2} \gets 1$ to $E_2$}{
    Sample a batch of data $B$ from $\mathcal{G}_{\text{train}}$ with batch size $B$\;
    \For {$G_j$, $y_j \in \mathcal{B}$}{
        Determine the environment of each sample $(G, y)$ in $G_j$ by $\text{argmax}_{\hat{e}} f(\hat{e}|s(G),y)$ in last hierarchy $K-1$\;
        Compute loss $\mathcal{L}_{\text{inv}}(F; G_j, \hat{e})$ according to Eq. 11\;
        Backpropagate and optimize $F$\;
    }
}
\textbf{Return} the final estimated model $F$\;
\label{algo:1}
\end{algorithm}

%------------------------------------------------------------------------
\begin{table*}[t]
\centering
% \fontsize{9}{9}\selectfont
\caption{\textbf{Statistics of the datasets used in experiments. The number of nodes is averaged numbers among all datasets.}}
\begin{tabular}{l|lllllll}
\toprule
\textbf{\textsc{Datasets}} & \textbf{\#\textsc{Training}} & \textbf{\#\textsc{Validation}} & \textbf{\#\textsc{Testing}} & \textbf{\#\textsc{Labels}} & \textbf{\#\textsc{Envs}} & \textbf{\#\textsc{Nodes}} & \textbf{\#\textsc{Metrics}}  \\ 
\midrule
\textsc{cminst-sp} & 40,000 & 5,000 & 15,000 & 2 & N/A & 56.90 & ACC \\
\textsc{graph-sst5} & 6,090 & 1,186 & 2,240  & 5 & N/A & 19.85 & ACC\\
\textsc{graph-twitter} & 3,238 & 694 & 1,509  & 3 & N/A & 21.10 & ACC\\
\textsc{ic50-assay} & 34,179 & 19,028 & 19,032 & 2 & 311 & 34.58 & ROC-AUC\\
% & 75.27 \\
\textsc{ic50-sca}   & 21,519 & 19,041 & 19,048 & 2 & 6,881 & 39.38 & ROC-AUC\\
% & 85.76 \\
\textsc{ic50-size}  & 36,597 & 17,660 & 16,415 & 2 & 190 & 37.99 & ROC-AUC\\ 
% & 82,78 \\
\textsc{ec50-assay} & 4,540 & 2,572 & 2,490 & 2 & 47 & 39.81 & ROC-AUC\\
% & 85.03 \\
\textsc{ec50-sca}   & 2,570 & 2,532 & 2,533 & 2 & 850 & 56.84 & ROC-AUC\\
% & 120.06 \\
\textsc{ec50-size}  & 4,684 & 2,313 & 2,398 & 2 & 167 & 48.40 & ROC-AUC\\
% & 103.06 \\
\bottomrule
\end{tabular}
\label{table_dataset}
\end{table*}

\section{Data Statistics}

\paragraph{DrugOOD datasets.} We present the data statistics for the DrugOOD dataset in \cref{table_dataset}. To evaluate the performance of our model in realistic scenarios involving distribution shifts, we have integrated three datasets from the DrugOOD benchmark \cite{ji2023drugood}. This benchmark offers a comprehensive out-of-distribution (OOD) evaluation in AI-driven drug discovery, specifically focusing on predicting drug-target binding affinity for both protein targets and drug compounds. The molecular data and annotations are precisely curated from the ChEMBL 29 database \cite{mendez2019chembl}, capturing various distribution shifts across various assays, scaffolds, and molecule sizes. We specifically select datasets with three different splits, assay, scaffold, and size, from the ligand-based affinity prediction task. This task involves IC50 and EC50 measurements and encompasses annotation noise at the core level.

\paragraph{General datasets.}
In \textit{CMNIST-sp} dataset, digits 0-4 are assigned to $y=0$, and digits 5-9 are assigned to $y=1$. Following previous studies, the label $y$ will be flipped with a probability of $0.25$. In addition, green and red colors will be assigned to images with labels $0$ and $1$ on an average probability of $0.15$ for the training data, respectively. In \textit{Graph-SST} datasets, the node features are generated using BERT \cite{kenton2019bert} and the edges are parsed by a Biaffine parser \cite{gardner2018allennlp}. We follow previous works \cite{chen2024does, chen2022learning} and split the datasets according to the averaged degrees of each graph. In the Graph-SST5 dataset, we partition graphs into the training and test sets using a sorting approach that arranges them from small to large degrees. In contrast, in the Twitter dataset, we adopt an inverse ordering strategy. This means that during training, the model is exposed to graphs with larger degrees, allowing us to assess its generalization performance on graphs with smaller degrees during evaluation.

%------------------------------------------------------------------------
 \section{Baselines and Experimental Settings}

\paragraph{GNN encoder.} For a fair comparison, we employ the same graph neural network (GNN) architecture in all methods. Consistent with prior studies, we utilize a multi-layer graph isomorphism network (GIN) \cite{xu2018powerful} with Batch Normalization \cite{ioffe2015batch} between layers and jumping knowledge (JK) residual connections at the last layer. The hidden dimension of the GNNs is set to 128, and we explore the number of GNN layers within the range of $\{3, 4, 5\}$.

\paragraph{Optimization and model selection.} We utilize the Adam optimizer \cite{kingma2014adam} and perform a grid search for learning rates in $\{1\mathrm{e}{-}3, 1\mathrm{e}{-}4, 4\mathrm{e}{-}5, 1\mathrm{e}{-}5\}$, as well as batch sizes in $\{32, 64, 128\}$ across all models and datasets. Early stopping based on validation performance is implemented with a patience of $20$ epochs. A default dropout value of 0.5 is used for all datasets. The final model is selected based on its performance on the validation set. All experiments are conducted with five different random seeds (1, 2, 3, 4, 5), and the results are reported as the mean and standard deviation from these five runs.

\paragraph{Euclidean OOD methods.} 
Building on the implementation of Euclidean OOD methods (IRM \cite{arjovsky2019invariant}, V-Rex \cite{krueger2021out}, IB-IRM \cite{ahuja2021invariance}, and EIIL \cite{creager2021environment}) in the previous study \cite{chen2024does}, we adopt the same implementation setup to reproduce the results of these methods. As environmental information is not provided, we follow the methodology of previous works \cite{chen2024does, chen2022learning} and randomly assign training data to two environments. The weights for the regularization term are selected from $\{1.0, 0.01, 0.001\}$ to calculate the IRM loss. Since the obtained results closely align with those reported in \cite{chen2024does}, we directly use the reported results from \cite{chen2024does} for comparison.

\begin{table}[t]
\centering
\begin{tabular}{llll}
\toprule
\textsc{methods} & \textsc{CMNIST-sp} & \textsc{Graph-SST5} & \textsc{Twitter} \\ 
\midrule
\textsc{erm} & 21.56$\pm$5.38 & 42.62$\pm$2.54 & 59.34$\pm$1.13\\
\midrule
% \textsc{irm}     & 23.29$\pm$7.82  & 42.77$\pm$1.26 & 60.42$\pm$1.06\\
% \textsc{v-rex}   & 24.62$\pm$8.75  & 42.48$\pm$1.67 & 60.50$\pm$2.05\\
% \textsc{eiil}    & 24.55$\pm$13.3  & 43.79$\pm$1.19 & 60.15$\pm$1.44\\
% \textsc{ib-irm}  & 13.06$\pm$1.97  & 43.02$\pm$1.94 & 60.80$\pm$2.50\\
% \midrule
\textsc{grea}    & 18.64$\pm$6.44 & 43.29$\pm$0.85 & 59.92$\pm$1.48\\
\textsc{disc}    & 54.07$\pm$15.3 & 40.67$\pm$1.19 & 57.89$\pm$2.02\\
\textsc{cigav1}  & 23.66$\pm$8.65 & 44.71$\pm$1.14 & 63.66$\pm$0.84\\
\textsc{cigav2}  & 44.91$\pm$4.31 & \underline{45.25$\pm$1.27} & \underline{64.45$\pm$1.99}\\
\textsc{gil}     & 18.04$\pm$4.39 & 43.30$\pm$1.24 & 61.78$\pm$1.66\\
\textsc{moleood} & 39.55$\pm$4.35 & 40.36$\pm$1.85 & 59.26$\pm$1.67\\
\textsc{gala}    & \textbf{59.16$\pm$3.64} & 44.88$\pm$1.02 & 62.45$\pm$0.62\\
\midrule
\rowcolor{Cyan!10} \textsc{ours} &  \underline{46.37$\pm$4.33} & \textbf{46.83$\pm$0.54} & \textbf{64.51$\pm$0.10} \\ 
\bottomrule
\end{tabular}
\caption{Test accuracies of various models on \textsc{CMNIST-sp}, \textsc{Graph-SST5}, and \textsc{Twitter} benchmark datasets. The mean $\pm$ standard deviation of all models is reported as an average of 5 executions of each model. The best performance in each dataset is highlighted in \textbf{bolded} and the second best methods are \underline{underlined}.}
\label{table_others}
\end{table}

\paragraph{Graph OOD methods.} We reproduce the experiments for GREA, CIGAv1, CIGAv2, MoleOOD, and GALA using the provided codes, if available. Although the previous work GALA \cite{chen2024does} implemented these baselines, our reproduced results differ. We adhere to the author-recommended hyperparameters for training the baselines, and consequently, some of the baselines exhibit superior results compared to the reported GALA results.
For the remaining baselines, including LiSA \cite{yu2023mind}, DisC \cite{fan2022debiasing}, and GIL \cite{li2022learning}, we observe that these models struggle to fit the DrugOOD dataset with the hyperparameters provided, as indicated in \cref{table_drugood_rest}.

\begin{figure}[t]
  \centering
  \captionsetup{font={scriptsize}}
  \includegraphics[width=1.0\linewidth]{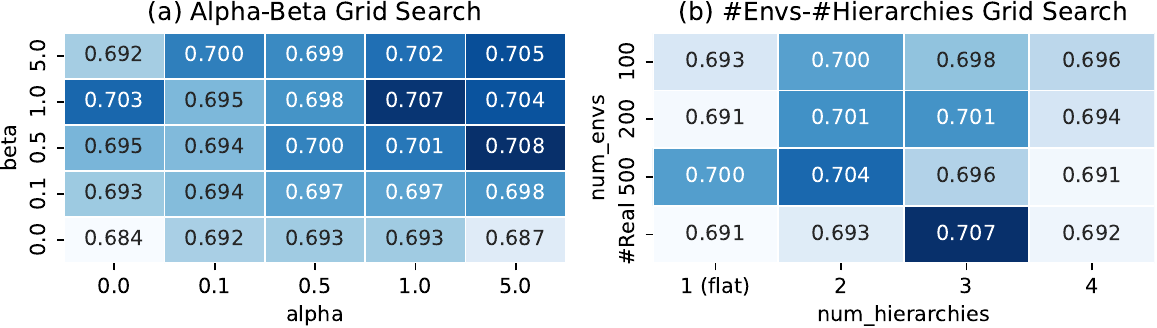}
   \caption{Hyperparameter Selection of alpha, beta, \#Envs, and \#Hierarchies.}
  \vspace{-5pt}
   \label{fig:rebuttal-hyper}
\end{figure}

\paragraph{Our methods.} As previously mentioned, we employ the same GNN architecture to obtain graph embeddings in our methods. In stochastic subgraph generation, we explore threshold values in $\{0.5, 0.6, 0.8\}$ for different datasets. To address potential over-smoothing issues in GNNs, we set the number of GNN layers in the environment inference stage (first stage) to 1 and search the number of GNN layers in graph invariant learning (second stage) in $\{3, 4, 5\}$.
For hierarchical settings, we search for the optimal number of hierarchy layers (i.e., $K$) from $\{1 ,2, 3\}$ for different datasets. As shown in \cref{table_dataset}, some datasets, such as CMNIST-sp and Graph-SST, do not provide environmental information. For general datasets, we set the number of environment labels in the first hierarchy (i.e., $E_1$) to 100. For DrugOOD datasets, we set $E_1$ as the provided number, e.g., $E_1$ of the \textsc{ic50-sca} dataset is 6881.
We search for the optimal number of environment labels at the last hierarchy in $\{2, 5, 10\}$ for different datasets. For intermediate hierarchies, we set the median as the number of environment labels, e.g., $K=3$ and $(E_1, E_2, E_3)=(100,50,2)$. We iteratively conduct hierarchical environment inference to obtain more high-level semantic environments.
In graph variant learning with inferred environments, we select the weights for the regularization term from $\{1.0, 0.01, 0.001\}$ to calculate the IRM loss.
% rebuttal contents 2024/03/04
In \cref{fig:rebuttal-hyper} (a), we conducted experiments for two inter-hierarchy contrastive learning. The performance of the model shows an upward trend with varying values of both $\alpha$ and $\beta$, demonstrating the effectiveness of incorporating $L_{EnvCon}$ and $L_{LabelCon}$ in our model. In \cref{fig:rebuttal-hyper} (b), the hierarchical inference of environments shows superiority over the flat inference (= existing methods) and the selection of the starting number of environments can have different performances.

\section{Additional Experiments}
We report results on general datasets (CMNIST-SP, Graph-SST5, and Twitter) as shown in \cref{table_drugood_rest}. In particular, our methods consistently outperform baseline approaches on Graph-SST datasets. We acknowledge that our methods did not yield successful results on the CMNIST-SP dataset.
We analyze the potential reasons as follows: (1) Our model focuses on learning variant substructures hierarchically, which can help infer more meaningful semantic environments for graph invariant learning.
(2) While CMNIST-SP contains shifts in node attributes, Graph-SST encounters shifts in node degree (structure). Our method demonstrates robustness in graphs featuring more complex distribution shifts, particularly those influenced by structural variations, as indicated by results on DrugOOD.
% rebuttal contents 2024/03/04
Furthermore, in \cref{rebuttal-new-ogb}, our method outperforms baseline performance when evaluated on three molecular property prediction datasets from the Open Graph Benchmark under fair conditions. As shown in \cref{rebuttal-new-baseline}, our model outperforms the additional three baselines on \textsc{ec50-sca} and \textsc{molhiv} datasets.
\begin{table}[t]
\centering
\begin{tabular}{l|lll}
\toprule
\textbf{\textsc{methods}}  & \textbf{\textsc{ec50-assay}} & \textbf{\textsc{ec50-sca}} & \textbf{\textsc{ec50-size}} \\ 
\textbf{\textsc{\#env}} & 47 & 850 & 167 \\ 
\midrule
\textsc{lisa \cite{yu2023mind}}     &  68.10$\pm$2.60 & 64.60$\pm$1.70 & 61.20$\pm$1.90  \\ 
% \midrule
\textsc{disc \cite{fan2022debiasing}}    & 65.40$\pm$5.34 & 54.97$\pm$3.86 & 56.97$\pm$2.56 \\
\textsc{gil \cite{li2022learning}}     & 72.13$\pm$4.70 & 63.05$\pm$1.04 & 62.08$\pm$1.06 \\
\midrule
\rowcolor{Cyan!10}\textsc{ours} &  \textbf{80.82$\pm$0.21} & \textbf{69.73$\pm$0.21} & \textbf{66.87$\pm$0.38} \\ 
\bottomrule
\end{tabular}
\caption{Test ROC-AUC of rest models on DrugOOD benchmark datasets. The mean $\pm$ standard deviation of all models is reported as an average of 5 executions of each model. The best methods are highlighted in \textbf{bold}.
}
\label{table_drugood_rest}
\end{table}

\begin{table}[t]\small
\renewcommand\arraystretch{0.5}
\centering
\captionsetup{font={small}}
\begin{tabular}{l|lll}
\toprule
\textsc{methods} & \textsc{molbace} & \textsc{molbbbp} & \textsc{molhiv}\\ \midrule
\textsc{gin} & 77.83\tiny{$\pm$3.15} & 66.93\tiny{$\pm$2.31} & 76.58\tiny{$\pm$1.02} \\
\textsc{gin+moleood} & 81.09\tiny{$\pm$2.03} & 69.84\tiny{$\pm$1.84} & 78.31\tiny{$\pm$0.24} \\
\midrule
\rowcolor{Cyan!10} \textbf{\textsc{ours}} & \textbf{82.26}\tiny{\textbf{$\pm$0.88}} & \textbf{71.30}\tiny{\textbf{$\pm$0.47}} & \textbf{79.65}\tiny{\textbf{$\pm$0.14}} \\
\bottomrule
\end{tabular}
\caption{Molecular property prediction from Open Graph Benchmark (OGB).}
\label{rebuttal-new-ogb}
\end{table}

\begin{table}[t]\small
\renewcommand\arraystretch{0.5}
\centering
\captionsetup{font={small}}
\begin{tabular}{l|llll}
\toprule
\textsc{datasets} & \textsc{dir} & \textsc{gsat} & \textsc{gil} & \cellcolor{Cyan!10} \textbf{\textsc{ours}}\\ \midrule
\textsc{ec50-sca} & 64.45\tiny{$\pm$1.69} & 66.02\tiny{$\pm$1.13} & 63.05\tiny{$\pm$1.04} & \cellcolor{Cyan!10} \textbf{69.73}\tiny{\textbf{$\pm$0.21}} \\
\textsc{molhiv} & 77.05\tiny{$\pm$0.57} & 76.47\tiny{$\pm$1.53} & 79.08\tiny{$\pm$0.54} & \cellcolor{Cyan!10} \textbf{79.65}\tiny{\textbf{$\pm$0.14}}\\
\bottomrule
\end{tabular}
\caption{More baseline comparisons in \textsc{ec50-sca} (DrugOOD) and \textsc{molhiv} (OGB).}
\label{rebuttal-new-baseline}
\end{table}

\section{Qualitative Analysis of Hierarchical Semantic Environments.}

\begin{figure*}
  \centering
  \begin{subfigure}{0.49\linewidth}
    % \fbox{\rule{0pt}{2in} \rule{.9\linewidth}{0pt}}
    \includegraphics[width=1.0\linewidth]{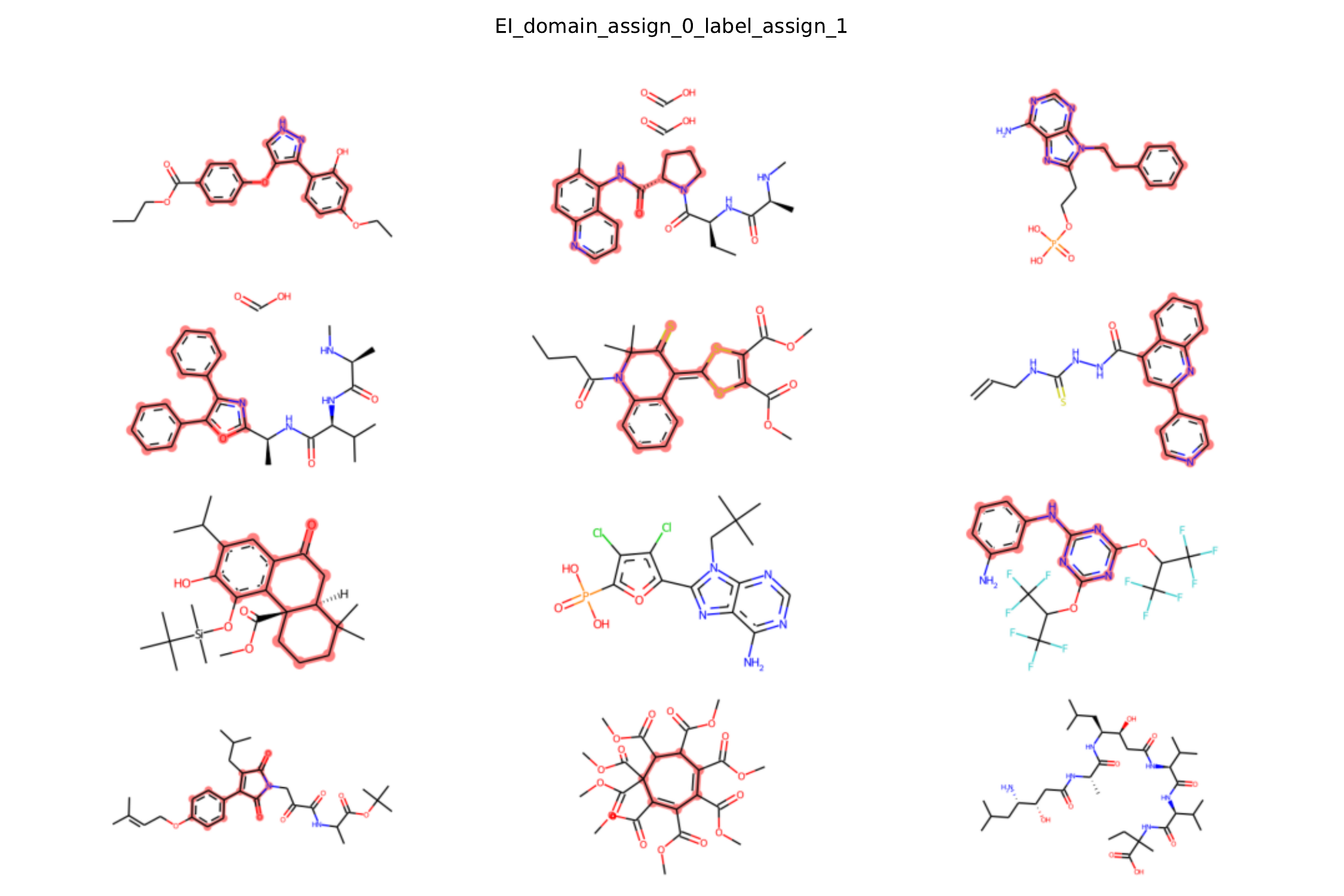}
    \caption{True label: 0; Predicted label: 1; Env: 0}
    \label{fig:EI-env-0}
  \end{subfigure}
  \hfill
  \begin{subfigure}{0.49\linewidth}
    % \fbox{\rule{0pt}{2in} \rule{.9\linewidth}{0pt}}
    \includegraphics[width=1.0\linewidth]{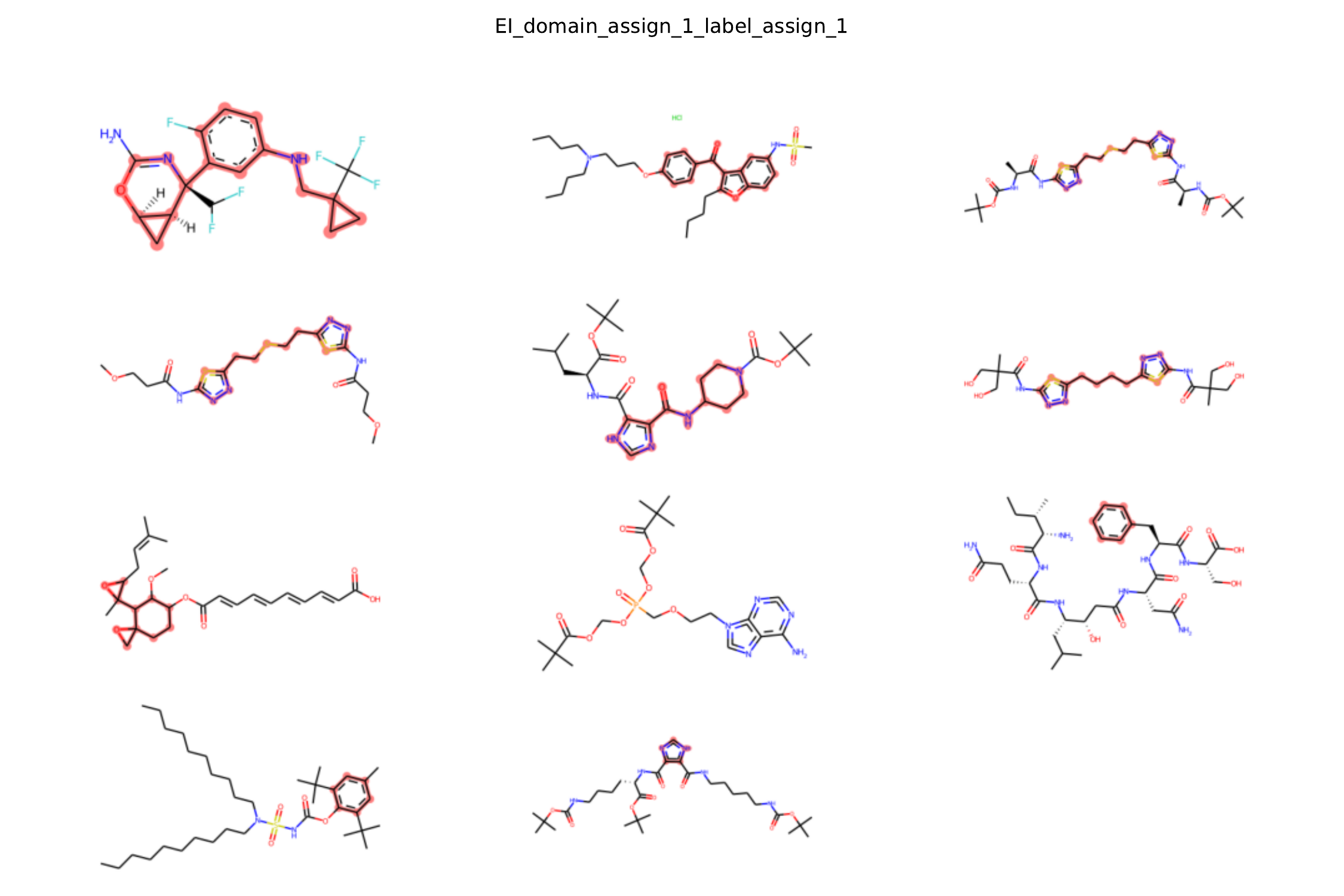}
    \caption{True label: 0; Predicted label: 1; Env: 2}
    \label{fig:EI-env-1}
  \end{subfigure}
  \caption{Visualizing molecules with incorrect predictions in their environments assigned by \textit{Infer\#2} model in \textsc{ic50-sca} dataset. The highlighted part with red color represents the scaffold information of each molecule.}
  \label{fig:EI-env}
\end{figure*}

\begin{figure*}
  \centering
  \begin{subfigure}{0.49\linewidth}
    % \fbox{\rule{0pt}{2in} \rule{.9\linewidth}{0pt}}
    \includegraphics[width=1.0\linewidth]{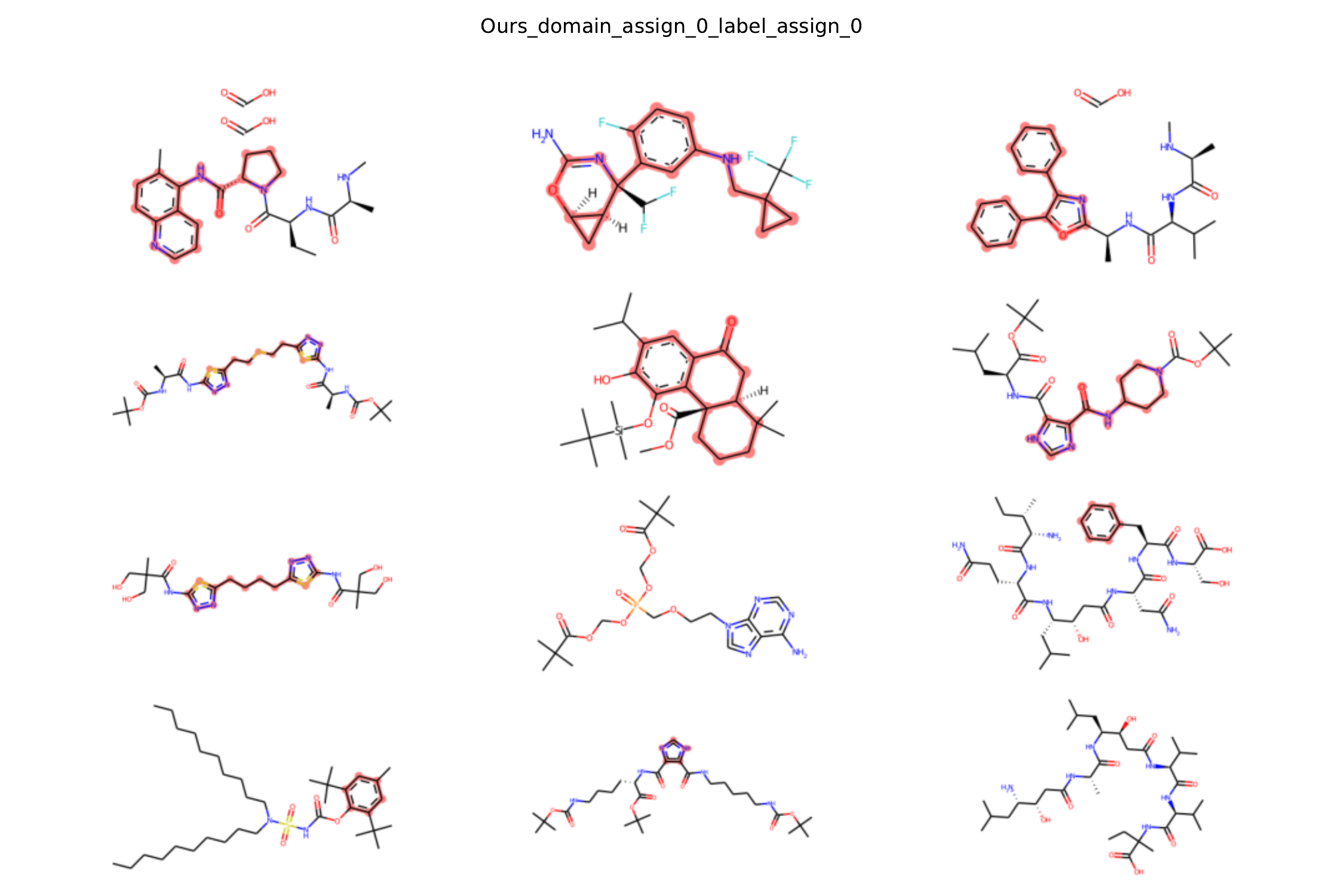}
    \caption{True label: 0; Predicted label: 0; Env: 0}
    \label{fig:our-env-0}
  \end{subfigure}
  \hfill
  \begin{subfigure}{0.49\linewidth}
    % \fbox{\rule{0pt}{2in} \rule{.9\linewidth}{0pt}}
    \includegraphics[width=1.0\linewidth]{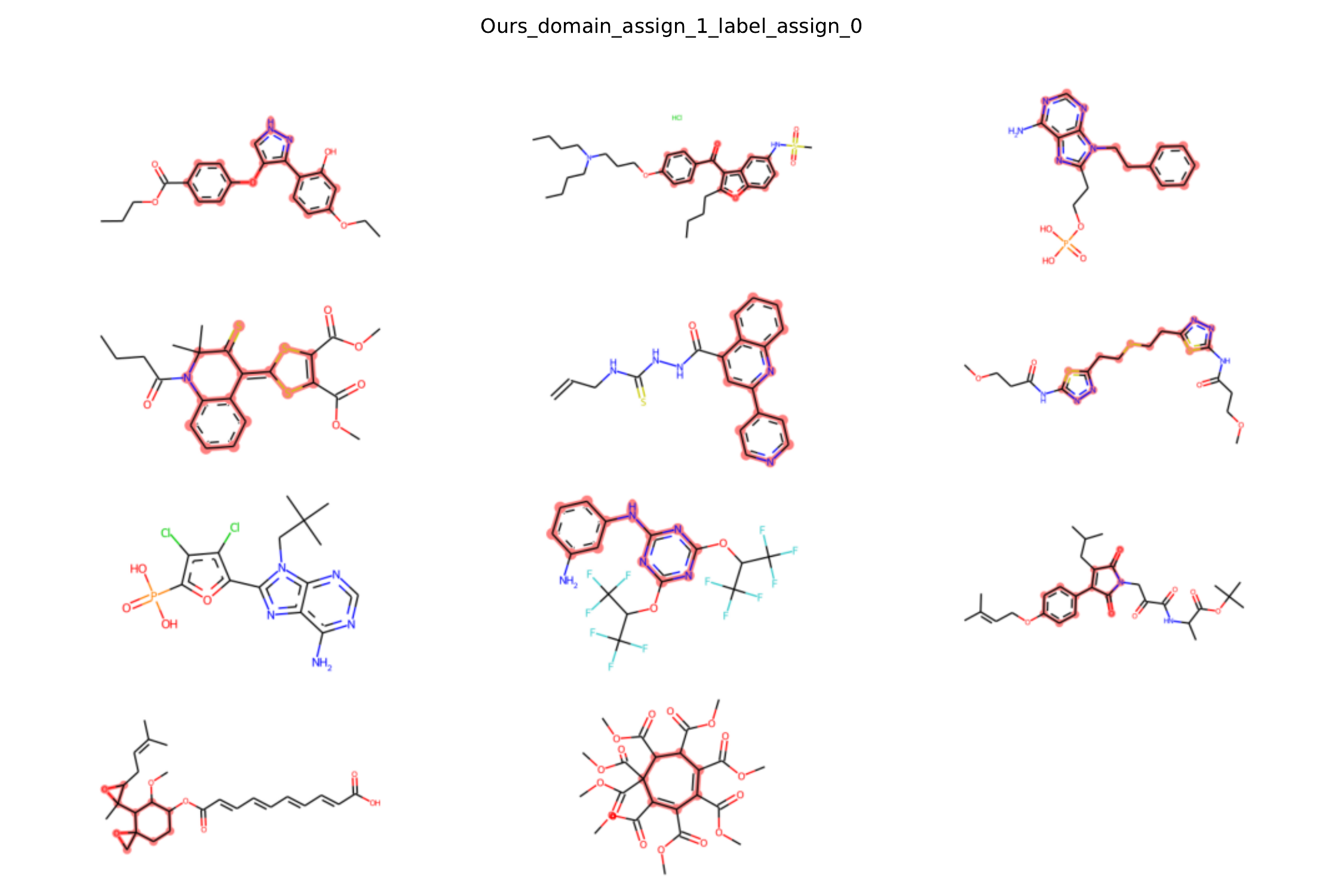}
    \caption{True label: 0; Predicted label: 0; Env: 1}
    \label{fig:our-env-1}
  \end{subfigure}
  \caption{Visualizing molecules with incorrect predictions in their environments assigned by our model in \textsc{ic50-sca} dataset. The highlighted part with red color represents the scaffold information of each molecule.}
  \label{fig:our-env}
\end{figure*}

To evaluate the practical efficacy of our proposed hierarchical environmental inference model in capturing complex environmental patterns, we utilize a visualization-based approach to compare the assignment outcomes between a flat environment model (\cref{fig:EI-env}) and our model (\cref{fig:our-env}) for the same set of molecules.
We specifically focus on molecules labeled 0 from the \textsc{ic50-sca} dataset, where our model exhibits accurate predictions while other flat environment-inference models (such as the \textit{Infer\#2} model) fail to predict accurately. We then visualize the allocation of these molecules in each environment, with particular emphasis on highlighting scaffold information.

In the \textit{Infer\#2} model, the necessity of directly predicting assignments to two environments in molecules with intricate scaffold structures may result in overfitting complex structural features.
In contrast, our model addresses the challenge of overfitting to molecular structural features. Through the combined learning of hierarchical environmental inference and graph invariant learning, we effectively extract spurious features, facilitating higher-level assignments of molecules to distinct environments.

In \cref{fig:EI-env}, the \textit{Infer\#2} model may allocate seemingly similar molecules to the same environment, but environments inferred from our model have a higher inter-environment diversification score, which is described in the discussion section (Sec.5.4). This indicates that the \textit{Infer\#2} model fails to capture spurious features, instead blending causal features and resulting in inaccurate predictions.
In summary, our hierarchical environmental assignment model generates more sophisticated semantic environments by learning hierarchical spurious features through graph invariant learning. This approach mitigates the overfitting challenges observed in models with flat environment inference, leading to a more accurate delineation of meaningful environments.
\begin{figure}[t]
  \centering
  \captionsetup{font={small}}
  \includegraphics[width=0.8\linewidth]{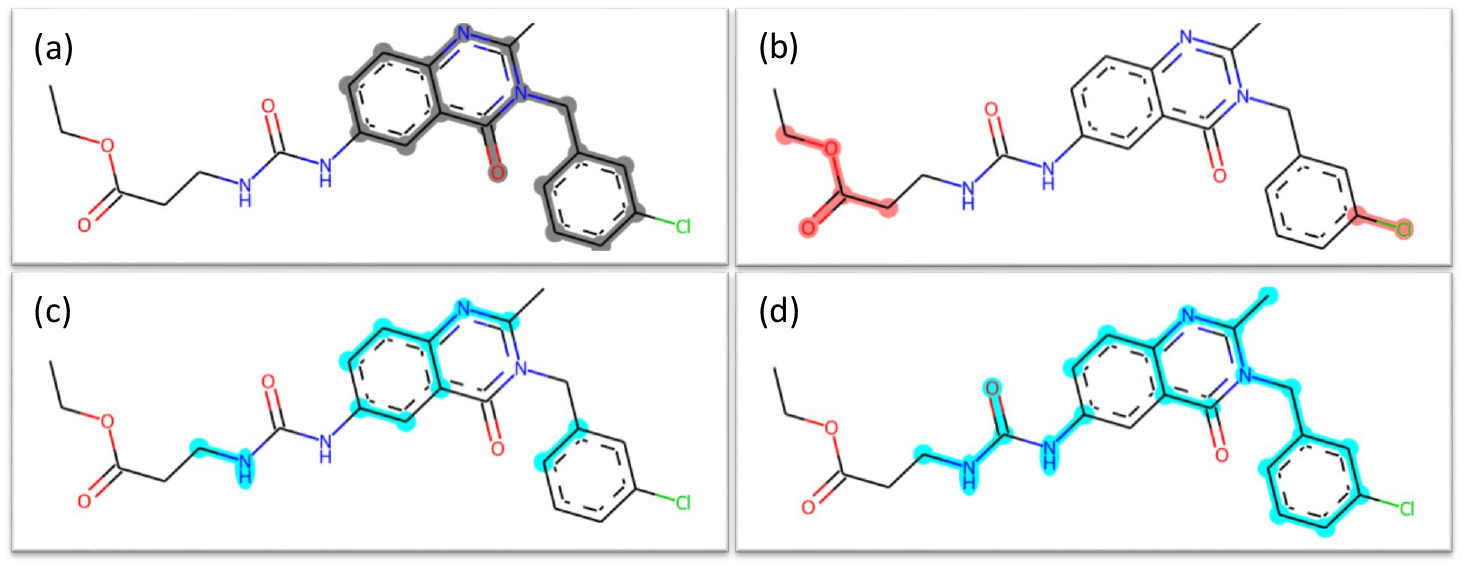}
   \caption{(a) Grey: Scaffold. (b) Red: Functional group (Chlorobenzene), and structural alert \textsc{COC(=O)C} from PubChem. (c, d) Blue: Learned variant subgraphs from the first and second hierarchies (the rest of the graph is considered learned invariant).}
   % \vspace{-10pt}
   \label{fig:rebuttal-case}
\end{figure}

\subsection{Specific Case Studies on Learned Invariant Graphs.}
In the DrugOOD-SCA dataset, consisting of 6,881 diverse scaffolds (environments), we computed Morgan-fingerprint Tanimoto similarity for all pairwise scaffolds. The results show a long-tailed distribution with a mean of 0.13 and a standard deviation of 0.04. This reveals the importance of learning the hierarchy of complex environments, given the presence of both similar and dissimilar scaffolds. We performed a specific case study using an active molecule from DrugOOD.
\cref{fig:rebuttal-case} shows our model capturing scaffold-like variants hierarchically while preserving invariant substructures. The final remaining invariant subgraphs are shown to be aligned with the functional group and structural alert.

\subsection{Theoretical Analysis of Our Method.}
Our method enhances out-of-distribution (OOD) performance by hierarchically stacking intra-hierarchy and inter-hierarchy losses.
Specifically, grounded in the theorem of \textit{Structrual Causal Models} \cite{arjovsky2019invariant}, the intra-hierarchy loss $L_{ED}$ satisfies the condition $H(Y|X_v) - H(Y|X_v, E_{\text{learn}}) > 0$ from EDNIL \cite{huang2022environment}, thereby maximizing the diversity of inferred environments at each hierarchy.
Furthermore, the $L_{\text{EnvCon}}$ and $L_{\text{LabelCon}}$ in the inter-hierarchy adjust the boundary of the embedding space for variant and invariant subgraphs, resulting in both theoretically and empirically diverse and reliable environments for graph invariant learning (\cref{fig:rebuttal-hyper}).

\end{document}